\documentclass{article} 
\usepackage[final]{colm2026_conference}

\usepackage{microtype}
\usepackage{hyperref}
\usepackage{url}
\usepackage{booktabs}

\usepackage{lineno}

\definecolor{darkblue}{rgb}{0, 0, 0.5}
\hypersetup{colorlinks=true, citecolor=darkblue, linkcolor=darkblue, urlcolor=darkblue}

\usepackage{latexsym}
\usepackage{adjustbox}
\usepackage{amsmath}
\usepackage{multirow}
\usepackage{graphicx}
\usepackage{caption}
\usepackage{stackengine}
\usepackage{listings}
\usepackage{comment}

\usepackage{tcolorbox}
\usepackage{xcolor}
\tcbuselibrary{breakable}

\usepackage{enumitem}
\usepackage{amssymb}
\usepackage{wrapfig}

\usepackage[table]{xcolor}
\usepackage[dvipsnames]{xcolor}

\newtcolorbox{configbox}{
  breakable,
  colback=gray!5,
  colframe=gray!60,
  boxrule=0.6pt,
  arc=3pt,
  left=2mm,
  right=2mm,
  top=1.5mm,
  bottom=1.5mm,
  fontupper=\small\ttfamily,
  title=Model Configuration,
  fonttitle=\bfseries,
}

\title{From Terminology to Diagrams: Visual-Instruction Generation for Scientific Diagram Understanding}

\author{Raul Ortega \& José Manuel Gómez-Pérez  \\
Language Technology Research Laboratory \\
Expert.ai\\
17 Henri Dunant, 28036 Madrid, Spain \\
\texttt{\{rortega, jmgomez\}@expert.ai} \\
}
\begin{document}

\ifcolmsubmission
\linenumbers
\fi

\maketitle

\begin{abstract}
Vision–language models (VLMs) have demonstrated strong performance in visual question answering with natural images. However, they continue to struggle with scientific diagrams, which are designed to convey functional or relational meaning rather than literal scenes. We therefore introduce a framework for generating large-scale diagram-grounded instruction data by leveraging terminology derived from scientific curricula. Our approach systematically extracts domain concepts, synthesizes atomic facts, retrieves relevant diagrams from the web, and generates multimodal supervision in the form of diagram captions and multiple-choice questions. Using this pipeline, we construct SciGram, a dataset of over 194K diagrams and 1.4M visual instructions across life, earth, and physical sciences. Despite relying on noisy web data and synthetic annotations, models fine-tuned on SciGram achieve substantial improvements on diagram-centric benchmarks, including TQA, ScienceQA, and AI2D, outperforming or matching state-of-the-art VLMs while using fewer training instances. Furthermore, augmenting existing models such as LLaVA OneVision with SciGram establishes new state-of-the-art performance on diagram question answering. Our results highlight the effectiveness of terminology-grounded instruction generation as a general strategy for improving vision-language reasoning in scientific domains.
To support future research in scientific diagram understanding, we release both the SciGram dataset and models.

\end{abstract}

\begin{figure*}[ht]
\centering
\includegraphics[width=0.95\linewidth]{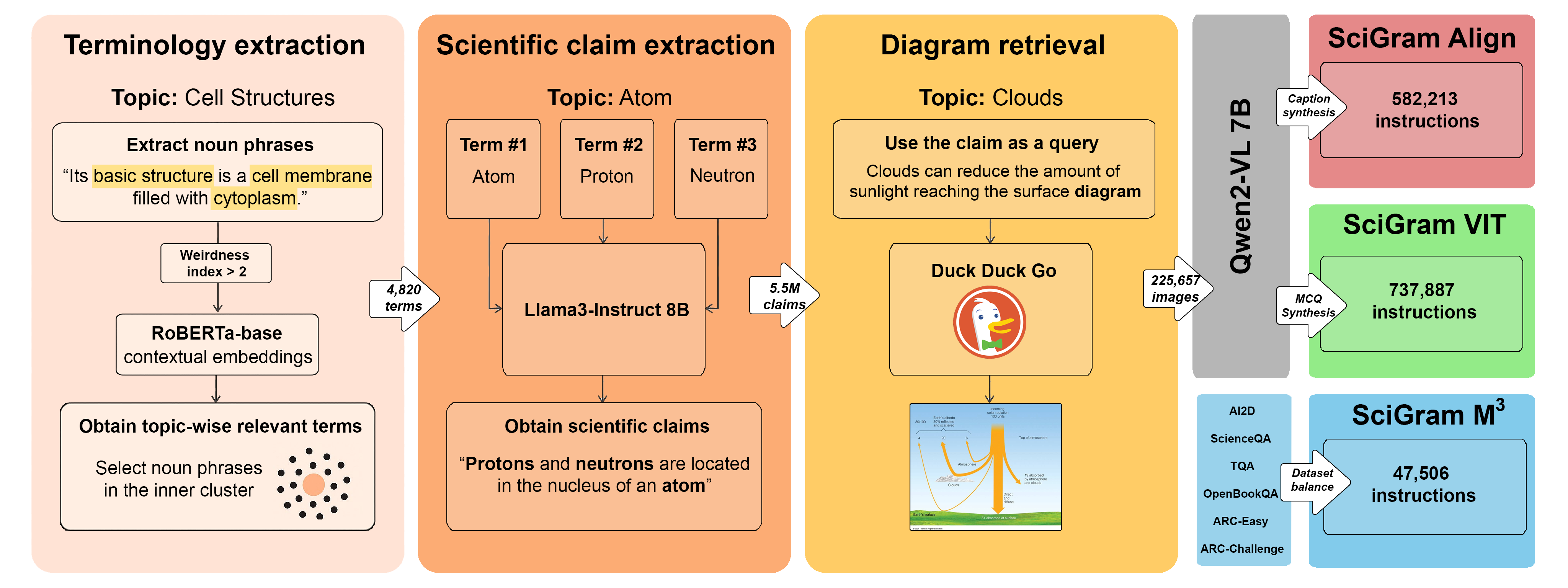}
\caption{Our six-stage dataset construction pipeline, comprising: terminology extraction, atomic fact generation, diagram retrieval, and SciGram subset generation (\textcolor{BrickRed}{\textbf{Align}}, \textcolor{ForestGreen}{\textbf{VIT}}, \textcolor{MidnightBlue}{\textbf{M\textsuperscript{3}}}).}
\label{fig:arch}
\end{figure*}

\section{Introduction}
\label{sec:intro}

In his 1988 AAAI Presidential Address, Raj Reddy identified a core \textit{AI Grand Challenge}: answering textbook-style questions requiring vision, language, reasoning, and learning~\citep{reddy_1988}. Today, this challenge is still largely unsolved in the natural sciences, where concepts like photosynthesis, the water cycle, and energy transfer combine textual explanations and supporting diagrams. Benchmarks such as AI2D~\citep{Kembhavi2016ADI}, TQA~\citep{Kembhavi_2017_CVPR}, and ScienceQA~\citep{lu2022learn} target this challenge by posing multimodal questions that require reasoning over scientific diagrams. However, despite advances in vision–language models (VLMs), scientific diagram understanding remains an open problem.

Scientific diagrams differ fundamentally from natural images: they are symbolic, abstract, and structurally diverse, conveying concepts, relationships, or processes rather than literal scenes~\citep{Kembhavi2016ADI}. Interpreting them requires grounding in scientific context, yet unlike natural images, scientific diagrams are scarce in existing training data for modern vision–language models. To address this gap, we propose a terminology-driven framework to create diagram-grounded instruction data for fine-tuning VLMs in scientific diagram understanding (see Figure~\ref{fig:arch}). This framework encompasses the extraction of concepts from middle-school science curricula, the generation of atomic scientific facts, retrieving their corresponding diagrams from the web, and synthesizing vision-language instructions grounded in those diagrams. Following this approach, we construct SciGram, a dataset of 194,071 scientific diagrams paired with synthetic instruction data. The main contributions of this work include the following:

\textbf{A terminology-driven framework} for constructing visual instruction datasets from scientific curricula and web data. 
Grounded in curriculum-derived terminology, our approach enables a broad coverage of relevant scientific concepts and vision-language supervision.

\textbf{The SciGram dataset}: Text–diagram pairs including captions and multiple-choice questions (MCQs) in the natural sciences (Figure~\ref{fig:scigram_example_lite}; additional examples in Appendix~\ref{sec:complete_examples}), 
in instruction-following format. Following large-scale VLM training trends, SciGram prioritizes coverage over precision, comprising over 194K Web diagrams and 1.4M synthetic instructions.

\textbf{The LLaVA-SciGram models}: A suite of VLMs built on the LLaVA architecture~\citep{llavapaper, 10655294} and fine-tuned on SciGram.

\textbf{A comprehensive evaluation}, showing that SciGram models outperform or match state-of-the-art VLMs and frontier models across scientific diagram understanding benchmarks. 


\begin{figure}[ht]
\centering

\begin{minipage}[t]{0.42\linewidth}
\vspace{0pt}
\centering
\includegraphics[width=\linewidth]{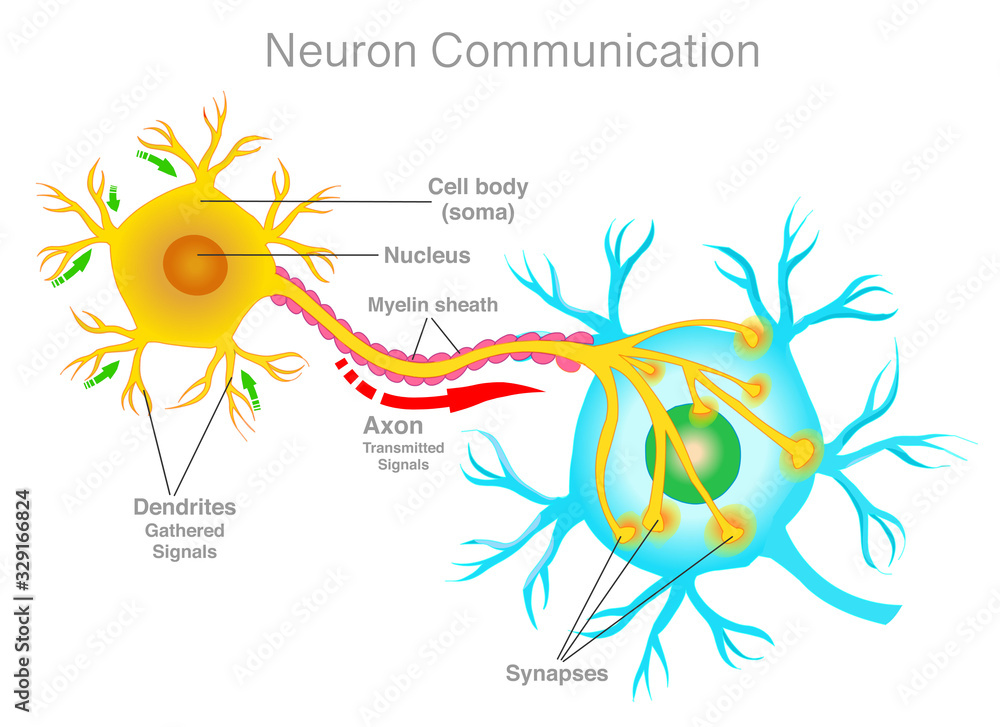}
\end{minipage}
\hfill
\begin{minipage}[t]{0.52\linewidth}
\vspace{0pt}

\tcbset{
    colback=BrickRed!5!white,
    colframe=BrickRed!75!black,
    boxrule=0.8pt,
    arc=4pt,
    left=2mm,
    right=2mm,
    top=1mm,
    bottom=1mm
}
\begin{tcolorbox}[title=SciGram-Align (Caption)]
\scriptsize
The diagram illustrates the process of neuron communication, showcasing the anatomy of two neurons and the flow of signals between them. On the left, a neuron with a yellow cell body (...)
\end{tcolorbox}

\tcbset{
    colback=ForestGreen!5!white,
    colframe=ForestGreen!75!black
}
\begin{tcolorbox}[title=SciGram-VIT (MCQ)]
\scriptsize
\textbf{What is the role of the myelin sheath?}\\
a) Protects the cell body \\
b) Carries signals from the cell body to the axon \\
c) Carries signals from dendrites to the cell body \\
d) Insulates the axon and speeds transmission $\checkmark$
\end{tcolorbox}

\end{minipage}

\caption{Example of SciGram diagram caption and multiple-choice question.}
\label{fig:scigram_example_lite}

\end{figure}

\section{Related work}
\label{sec:back}

Early work on scientific diagram understanding~\citep{Kembhavi_2017_CVPR} explored approaches from machine reading comprehension~\citep{Seo2017BidirectionalAF, Weston2014MemoryN}, visual question answering methods~\citep{Antol2015VQAVQ}, and diagram-specific parsers~\citep{Kembhavi2016ADI}, highlighting challenges distinct from natural images. Subsequent approaches included reasoning-focused models~\citep{li2018} and graph-based models~\citep{kim-etal-2019-textbook, ma2021, 10656098} to capture spatial and semantic relations. Transformer-based models, such as BERT~\citep{Devlin2018}, RoBERTa~\citep{liu2019robertarobustlyoptimizedbert}, and PaLM~\citep{chowdhery2022}, were extended to multimodal tasks, producing models like VL-BERT~\citep{su2019vl} and LXMERT~\citep{tan-bansal-2019-lxmert}. However, those early approaches focused exclusively on natural images. ISAAQ~\citep{gomez-perez-ortega-2020-isaaq} partially addressed this gap with cross-modal attention for diagram-based question answering.

Contrastive methods like CLIP~\citep{pmlr-v139-radford21a} and SIGLIP~\citep{zhai2023sigmoidlosslanguageimage} advanced pretraining by aligning image–text embeddings, forming the backbone of modern VLMs like LLaVA and MOLMo~\citep{deitke2024molmopixmoopenweights}, which combine visual encoders with large language models (LLMs). However, they rely on general-purpose instruction datasets, such as LLaVA OneVision~\citep{li2025llavaonevision} and PixMo~\citep{deitke2024molmopixmoopenweights}, with sparse coverage of scientific diagrams. In contrast, domain-specific VLMs like LLaVA-Med~\citep{li2023llavamedtraininglargelanguageandvision}, LLaVA-Chef~\citep{Mohbat2024LLaVAChef}, and LLaVA-Ultra~\citep{Guo2024LLaVAUltra} show the benefits of fine-tuning on specialized datasets. 
While datasets such as MMMU~\citep{yue2024mmmumassivemultidisciplinemultimodal}, VQA Abstract Scenes~\citep{Antol2015VQAVQ}, and SciVerse~\citep{guo2025sciverseunveilingknowledgecomprehension} contain diagrammatic images, they differ from the scientific diagrams represented in benchmarks such as AI2D, TQA, and ScienceQA (SQA), which visually illustrate specific scientific concepts. However, these benchmarks provide insufficient training data to effectively develop diagram reasoning, limiting current VLMs' ability to understand scientific content.


\section{Method}
\label{sec:app}

We propose a framework for generating multimodal instruction data for scientific diagram understanding, grounded in curriculum-derived scientific terminology to ensure broad domain coverage and semantic alignment between text and visual content. Unlike prior pipelines based primarily on free-form web data or captions, our approach follows a structured progression from terminology to instructions through four stages: i) terminology extraction, ii) atomic fact generation, iii) diagram retrieval, and iv) instruction generation. The resulting three complementary datasets align with VLM training pipelines such as LLaVA, which combine vision–language alignment with visual instruction tuning. Prompt templates and examples are provided in Appendices~\ref{sec:prompts} and~\ref{sec:instruction_examples}.

\subsection{Terminology extraction}

We begin by extracting scientific terminology from structured educational sources to provide a compact yet comprehensive set of domain concepts. We construct a terminology set covering middle-school natural sciences by leveraging the textbook used in~\citet{Kembhavi_2017_CVPR}. Following its organization by topics, we analyze each section, including lessons, explanations, and instructional materials, to identify terms linked to distinct semantic concepts. This serves as the semantic backbone of our data generation process, ensuring that all downstream steps remain grounded in meaningful scientific concepts rather than arbitrary web content. This process consists of three steps:

\textbf{Tokenization and noun-phrase identification.} For each topic $d$, we tokenize the text, discard stop words, and extract noun phrases, capturing a broader set of scientific concepts than named entities only. These noun phrases constitute a first set of term candidates $T_d$.

\textbf{Selection of distinctive terms.} To retain only domain-relevant noun phrases, we compute their weirdness index~\citep{conf_trec_1999}, which compares textbook term frequencies to a general corpus BNC\footnote{British National Corpus (\url{https://www.english-corpora.org/bnc})}. This value identifies terms that are characteristic of the target domain while reducing the influence of general-purpose vocabulary. Terms whose score exceeds a threshold $t$ are kept and lemmatized to merge morphological variants. We empirically set $t=2$ to filter out general and non-scientific terms from $T_d$.

\textbf{Embedding representation and clustering.} To obtain a semantically coherent set of domain terms, we embed each $t_i \in T_d$ using RoBERTa-base~\citep{liu2019robertarobustlyoptimizedbert}, a model which provides a computationally efficient and well-established semantic representation baseline for clustering and similarity filtering. We average the contextual representations of each term ($e_i$) across all sentences in which it appears. We then compute the centroid $c_d$ of these embeddings and measure Euclidean distances $\delta_i = |\mathbf{e}_i - \mathbf{c}_d|_2$, discarding terms beyond one standard deviation. We use Euclidean distance instead of cosine similarity to better capture the absolute scale of variation in the embedding space. 

This process yields a curated vocabulary of 4,820 distinct, semantically coherent scientific terms. Additional statistics on the selected terminology are provided in Appendix~\ref{sec:terminology_stats}.

\subsection{Atomic fact generation}

We generate atomic science facts for each textbook topic from its curated terminology to capture elementary scientific relationships. These facts serve as intermediate representations bridging concepts and visual grounding, providing fine-grained semantic anchors for retrieving relevant diagrams. 

For each topic $d$, we consider all non-empty combinations of its terminology, $C_d = \mathcal{P}(T_d) \setminus {\emptyset}$, where $\mathcal{P}(T_d)$ denotes the power set of $T_d$. We instruct a LLaMA3-8B-Instruct~\citep{grattafiori2024llama3herdmodels} to generate concise, factual, middle-school–level statements that include each combination $c \in C_d$, such as “Protons and neutrons are located in the nucleus of an atom”. To ensure broad coverage of concept interactions, the model is asked to produce up to 50 such statements for every combination. 

After deduplication, this process yields 5,508,218 unique facts across all topics.

\subsection{Diagram retrieval}

For each synthesized fact, we retrieve candidate images from the web and filter them to retain diagram-like content. We query DuckDuckGo\footnote{\url{https://pypi.org/project/duckduckgo-search}} using each atomic fact appended with the suffix "diagram" (e.g., “pollution affects human health, cognitive development, and immune systems diagram”), and collect the top five image results with URLs and metadata. This process ran 
on two cloud instances (4 CPUs and 16 GB RAM each) for 21 days.

To mitigate potential noise from web retrieval, such as natural images, irrelevant visuals, and stylistic artifacts, we apply several filtering steps: we retain only images linked to at least five atomic facts, ensuring that each diagram has sufficient textual support; remove duplicates via perceptual hashing~\citep{NIST_SHS}; and discard invalid or unsupported files
. Filtering is intentionally light to preserve scale. We accept residual noise as a trade-off for broad coverage, consistent with large-scale dataset construction practices~\citep{pmlr-v139-radford21a, li2025llavaonevision}. While this does not guarantee perfect scientific correctness, our consistent improvements across multiple benchmarks suggest that the resulting supervision signal is nevertheless effective for improving scientific diagram understanding. This process yields 255,657 unique images. 

\subsection{Instruction-following data generation}
\label{sec:instr_gen}

Given the filtered list of diagrams, we generate multimodal instruction data consisting of captions (diagram descriptions) and MCQs grounded in visual content. This dual supervision enables both descriptive and reasoning capabilities of models~\citep{llavapaper}. To this end, and given our hardware constraints, we use Qwen2-VL-7B~\citep{yang2024qwen2technicalreport}, a VLM that we found capable of producing reasonably detailed and context-aware textual descriptions from images.

\textbf{Caption synthesis.} We generate descriptive captions for each diagram to align textual and visual features. The model is instructed to generate a paragraph-form caption emphasizing key components, their relationships, and relevant spatial, temporal, or dynamic aspects. To increase diversity and reduce potential bias, we repeat the captioning process three times for each diagram. Using normalized Levenshtein similarity~\citep{levenshtein_binary_1966}, the average similarity score between captions for the same diagram is 0.4196, indicating substantial variation. These image-caption pairs are formatted into instruction-following examples using a naïve expansion strategy similar to the one proposed in~\citet{llavapaper}. The resulting dataset forms the alignment subset, which we refer to as \textcolor{BrickRed}{\textbf{SciGram-Align}}. 

\textbf{Multiple-choice question synthesis.} We generate diagram-grounded MCQs to create instruction-following data for reasoning over scientific diagrams. For each diagram, we instruct the model to produce MCQs relying solely on visual elements, phrased at a middle-school level, and covering domains across natural sciences. Duplicated questions (5.7\%) are discarded, and the distribution of correct answers is balanced across the four answer options. Questions are converted to a JSON instruction-following format, e.g., \{"answer": "b"\}, enabling consistent training and evaluation. This forms the \textcolor{ForestGreen}{\textbf{SciGram-VIT}} subset.

\textbf{Curation of existing datasets.} To provide a final stage of high-quality, domain-focused training aligned with our target tasks, we build the \textcolor{MidnightBlue}{\textbf{SciGram-M\textsuperscript{3}}} subset using diagram-based QA datasets (SQA, AI2D, TQA) present in the LLaVA OV training mixture, together with selected text-only QA sets (ARC-Easy/Challenge~\citep{Clark2018ThinkYH} and OpenBookQA~\citep{mihaylov-etal-2018-suit}). All questions are converted to instruction-following format, and answer choices are shuffled to reduce imbalance and overfitting (details in Appendix~\ref{sec:balancing}).

\section{The SciGram dataset}
\label{sec:scigram}

\subsection{Dataset structure}

The SciGram dataset consists of three subsets, \textcolor{BrickRed}{\textbf{SciGram-Align}}, \textcolor{ForestGreen}{\textbf{SciGram-VIT}} and \textcolor{MidnightBlue}{\textbf{SciGram-M\textsuperscript{3}}}, designed to be used at different stages of the training pipeline. Focused on scientific diagram understanding, SciGram is much more compact (1.4M instructions) than general-purpose alternatives such as the LLaVA OV data (7.8M).\footnote{\url{https://huggingface.co/datasets/lmms-lab/LLaVA-OneVision-Data}}

\textbf{\textcolor{BrickRed}{SciGram-Align}} contains 582,213 instruction pairs designed to align visual and textual features during the initial training stage via a captioning task. Each diagram is associated with three one-paragraph captions that provide detailed descriptions of the entities and processes in the image.

\textbf{\textcolor{ForestGreen}{SciGram-VIT}} consists of 737,887 instructions created to fine-tune the model on a multiple-choice question answering (MCQA) task involving diagrams. Each question has four answer options, with only one correct answer per question.

\textbf{\textcolor{MidnightBlue}{SciGram-M\textsuperscript{3}}} consists of 47,506 instructions from the training sets of TQA (14,050 questions), SQA (12,726), OpenBookQA (4,957), and ARC-Easy/Challenge (3,370). Since AI2D does not provide official splits, we used the same 12,403 questions as in LLaVA OV data.

\subsection{Human evaluation of dataset quality}
\label{sec:base_human_eval}

To assess data quality, four domain experts independently reviewed a random sample of 600 SciGram items, evenly distributed across diagrams, diagram-caption pairs, and multimodal MCQs. While raw inter-rater agreement is relatively high (82.41\%), Cohen’s $\kappa$ is low (0.27), as expected under strong class prevalence imbalance~\citep{DERKSEN2024482}. We therefore report Gwet’s AC1~\citep{Gwet2008-fu} as a more robust measure, yielding an average AC1 score of 0.59, indicating moderate-to-substantial agreement
. Detailed results are in Appendix~\ref{sec:human_eval}.

As expected for a web-crawled dataset, some noise is present, according to our annotators: 24\% of the retrieved images are not actual diagrams but natural images, charts, and others. Despite this, 88\% captions align with their diagrams, 82\% cover key elements and relations, 82\% match middle-school complexity, and 75\% provide interpretative value, i.e., they help the reader understand or reason about the diagram, rather than just describe it. For MCQs, 89\% are visually-grounded, 76\% match domain and difficulty, and 93\% are considered unambiguous and clearly phrased, with effective (89\%) and distinctive (92\%) distractors. 

The evaluation reveals other limitations: 61\% of MCQs may be answered using prior knowledge, potentially reducing diagram reliance, while 16\% show labeling inconsistencies, e.g., correct options marked as incorrect and vice versa, calling for stronger future verification procedures such as diagram/non-diagram classifiers and automated consistency verification models. Nevertheless, as Section \ref{sec:main_results} shows, fine-tuning on SciGram reports considerable benefits over the baselines. 

\section{Experimental Setup}

We evaluate the impact of SciGram on scientific diagram understanding using the LLaVA architecture, chosen for its modular vision–language alignment, instruction-tuning pipeline, and open-source availability
. We consider two regimes: (i) training from scratch with SciGram data at each stage, and (ii) fine-tuning a pretrained LLaVA-OV 7B model. The resulting models, LLaVA-SciGram 7B and LLaVA-SciGram OV 7B, are trained on two NVIDIA A100 GPUs, requiring approximately 450 GPU-hours each.

\textbf{LLaVA-SciGram 7B} consists of a pretrained CLIP vision encoder and Qwen2-Instruct 7B~\citep{yang2024qwen2technicalreport} as the language backbone. For training, we follow the same pipeline as LLaVA OV. First, an alignment stage in which visual features are aligned with the pretrained LLM embedding space. We train the projection matrix on \textcolor{BrickRed}{\textbf{SciGram-Align}}, keeping both the visual encoder and LLM weights frozen, for one epoch with a learning rate of 1e-3. Then, an instruction tuning stage using LoRA~\citep{hu2021loralowrankadaptationlarge} to train on \textcolor{ForestGreen}{\textbf{SciGram-VIT}} for one epoch with a learning rate of 1e-5. After merging the LoRA adapter into the model, we fine-tune another LoRA adapter on \textcolor{MidnightBlue}{\textbf{SciGram-M\textsuperscript{3}}} for 3 epochs with learning rate 1e-5. Additional details regarding hyperparameters are provided in Appendix~\ref{sec:hyperparameters}.

\textbf{LLaVA-SciGram OV 7B} follows the same fine-tuning but uses the pretrained weights of LLaVA OV trained on single images, with a SIGLIP vision encoder and Qwen2-Instruct 7B.

We evaluate LLaVA-based models fine-tuned on SciGram using three complementary diagram MCQA benchmarks: TQA, SQA, and AI2D. These datasets were selected to capture different aspects of scientific diagram understanding across grade levels, modalities, and reasoning types. To prevent data contamination from web-sourced images, all benchmark test diagrams are excluded from SciGram.

\textbf{TQA} contains text-only multiple-choice and true/false questions, as well as diagram-grounded questions. It covers physical, life, and earth sciences, using a text fragment or diagram as context; for questions without diagrams, the associated lesson serves as context. 

\textbf{SQA}  is collected from elementary and high school science curricula, and contains multimodal MCQs that can include diagram questions, text-only questions, and also questions with natural images, providing a broader coverage of modalities in the scientific domain.

\textbf{AI2D} contains grade-school science diagrams paired with MCQs. We use the most common split
\footnote{\url{https://huggingface.co/datasets/lmms-lab/ai2d}}, where diagrams have masked labels, requiring models to infer elements and processes visually. We also evaluate a variant with visible labels for a less challenging setting\footnote{\url{https://huggingface.co/datasets/lmms-lab/ai2d-no-mask}}.

\section{Results}
\label{sec:main_results}

We first assess SciGram fine-tuning on three benchmarks, then compare against diverse baselines, including larger non-LLaVA architectures. We next ablate each SciGram subset. Finally, using 200 randomly sampled TQA diagram-question (DQ) items, we analyze performance by knowledge/reasoning type and test visual–language integration on questions requiring diagram understanding.

\subsection{Effect of SciGram in LLaVA models}

Table~\ref{table:Allresults} compares the best-performing 7B LLaVA model (LLaVA OV) with LLaVA-SciGram 7B and LLaVA-SciGram OV 7B. Fine-tuning with SciGram consistently boosts performance, with gains up to 16 points. LLaVA-SciGram OV generally achieves the largest improvements, except on \textit{Language} and \textit{No Support} SQA questions, where LLaVA-SciGram 7B slightly outperforms. These results show that SciGram fine-tuning significantly enhances scientific QA, especially for questions involving visual understanding (TQA DQ, SQA IMG, AI2D).

\begin{table}[ht]
\small
\centering
\begin{tabular}{llccc}
 Dataset & Question type & LLaVA OV (Acc.)
 & LLaVA-SciGram ($\Delta$)
 & LLaVA-SciGram OV ($\Delta$)\\\midrule
\multirow{3}[0]{*}{TQA} & Text-only & 89.09 & \cellcolor{green!11}+1.78 & \cellcolor{green!12}+1.85 \\
& True/False & 88.49 & \cellcolor{green!22}+3.51 & \cellcolor{green!25}+3.94 \\
& Diagram & 77.08 & \cellcolor{red!3}-0.40 & \cellcolor{green!19}+3.04\\
& \textbf{Overall} & \textbf{82.70} & \cellcolor{green!6}\textbf{+0.96} & \cellcolor{green!18}\textbf{+2.87} \\ \midrule
& Natural Sciences & 88.10 & \cellcolor{green!51}+8.17 & \cellcolor{green!58}+9.24 \\
& Social Sciences & 88.98 & \cellcolor{green!53}+8.55 & \cellcolor{green!63}+10.12 \\
& Language & 78.64 & \cellcolor{green!81}+13.00 & \cellcolor{green!77}+12.27 \\
& Text Support & 92.40 & \cellcolor{green!42}+6.71 & \cellcolor{green!42}+6.71 \\
\multirow{2}[0]{*}{SQA} & Visual Support & 87.31 & \cellcolor{green!50}+7.93 & \cellcolor{green!64}+10.31 \\
& No Support & 80.14 & \cellcolor{green!83}+13.24 & \cellcolor{green!75}11.99 \\
& Grade 1-6 & 88.95 & \cellcolor{green!41}+6.54 & \cellcolor{green!45}+7.20 \\
& Grade 7-12 & 80.22 & \cellcolor{green!93}+14.84 & \cellcolor{green!98}+15.63 \\
& \textbf{Overall} & \textbf{85.83} & \cellcolor{green!59}\textbf{+9.50} & \cellcolor{green!64}\textbf{+10.21} \\ \midrule
\multirow{2}[0]{*}{AI2D} & Opaque labels & 79.50 & \cellcolor{green!4}+0.71 & \cellcolor{green!25}+3.95 \\
& Transparent labels & 89.90 & +0.03 & \cellcolor{green!16}+2.52 \\
& \textbf{Overall} & \textbf{84.70} & \cellcolor{green!2}\textbf{+0.37} & \cellcolor{green!20}\textbf{+3.24} \\ \bottomrule
\end{tabular}
\caption{LLaVA variants performance with/without SciGram fine-tuning across benchmarks.}
\label{table:Allresults}
\end{table}

\subsection{Comparison with other models}

To contextualize SciGram's improvements, we compare our models against: i) prior state-of-the-art baselines; ii) recent multimodal models of similar size (Phi-3 Vision~\citep{abdin2024phi3technicalreporthighly}, MOLMo 7B, Pixtral 12B~\citep{agrawal2024pixtral12b}, Qwen2-VL 7B); iii) frontier multimodal models (Gemini 2.0 Flash~\citep{geminiteam2025geminifamilyhighlycapable}, GPT4o~\citep{openai2024gpt4technicalreport}); and iv) other LLaVA variants. To reduce bias from potential pretraining-test overlap in frontier models, answer options are shuffled. Results are partly reproduced from prior work and partly obtained with our evaluation pipeline (see Appendix~\ref{sec:evaluation_details}).

\begin{table}[ht]
\small
    \centering
    \begin{tabular}{lcccc}
    Model & Text MC & True/False & Diagram MC & Overall \\ \midrule
    Random & 22.88 & 50.10 & 24.96 & 29.08 \\
    MemN+VQA~\citep{Antol2015VQAVQ} & 31.05 & 50.50 & 31.82 & 35.11 \\
    MemN+DPG~\citep{Kembhavi_2017_CVPR} & 30.98 & 50.50 & 32.83 & 35.62 \\
    BiDAF+DPG~\citep{Seo2017BidirectionalAF} & 30.46 & 50.40 & 32.72 & 35.39 \\
    FCC+Vecsigrafo~\citep{10.1145/3360901.3364420} & 36.56 & - & 35.30 &  -\\
    IGMN~\citep{li2018} & 40.00 & 57.41 & 36.35 & 41.36 \\
    f-GCN1+SSOC~\citep{kim-etal-2019-textbook} & 49.54 & 62.73 & 37.61 & 45.77\\
    ISAAQ~\citep{gomez-perez-ortega-2020-isaaq} & 72.06 & 78.83 & 55.12 & 64.66 \\ \midrule
    Phi-3 Vision~\citep{abdin2024phi3technicalreporthighly} & 81.19 & 74.01 & 72.91 & 75.52 \\
    MOLMo 7B-D~\citep{deitke2024molmopixmoopenweights} & 81.35 & 84.80 & 71.20 & 76.69 \\
    Pixtral 12B~\citep{agrawal2024pixtral12b} & 85.84 & 91.85 & 77.08 & 82.39 \\
    Qwen2-VL 7B~\citep{yang2024qwen2technicalreport} & 87.69 & 91.63 & \underline{78.08} & 83.41 \\  \midrule
    Gemini 2.0 Flash~\citep{geminiteam2025geminifamilyhighlycapable} & \underline{90.94} & \underline{94.73} & 68.18 & 79.73\\
    GPT4o~\citep{openai2024gpt4technicalreport} & \textbf{94.20} & \textbf{96.16} & 77.32 & \textbf{85.74}\\ \midrule
    LLaVa 1.5~\citep{llavapaper} & 67.03 & 60.02 & 39.85 & 51.47 \\
    LLaVA OneVision 7B~\citep{li2025llavaonevision} & 89.09 & 88.49 & 77.08 & 82.70 \\
    LLaVA-SciGram 7B & 90.87 & 92.00 & 76.68 & 83.66 \\
    LLaVA-SciGram OneVision 7B & \underline{90.94} & 92.43 & \textbf{80.12} & \underline{85.57} \\
    \bottomrule
    \end{tabular}
    \caption{Performance comparison on the TQA test set. Values denote accuracy (\%). Bold indicates the best result; underlined indicates the second best.}
    \label{table:TQAresults}
\end{table}

As shown in Table~\ref{table:TQAresults}, LLaVA-SciGram OV establishes a new state of the art on TQA diagram questions, outperforming all baselines on the Diagram Multiple Choice (DMC) subset
. GPT4o achieves the highest overall accuracy, driven largely by text-only questions.

\begin{table}[ht]
\centering
\small
\begin{adjustbox}{width=\textwidth}
\begin{tabular}{lccccccccc}
Model & NAT & SOC & LAN & TXT & IMG & NO & G1-6 & G7-12 & Overall \\ \midrule
Human & 90.23 & 84.97 & 87.48 & 89.60 & 87.50 & 88.10 & 91.59 & 82.42 & 88.40\\ \midrule
MCAN \citep{yu2019deepmodularcoattentionnetworks} & 56.08 & 46.23 & 58.09 & 59.43 & 51.17 & 55.40 & 51.65 & 59.72 & 54.54\\
Top-Down \citep{anderson2018bottomup} & 59.50 & 54.33 & 61.82 & 62.90 & 54.88 & 59.79 & 57.27 & 62.16 & 59.02\\
BAN \citep{10.5555/3326943.3327087} & 60.88 & 46.57 & 66.64 & 62.61 & 52.60 & 65.51 & 56.83 & 63.94 & 59.37\\
DFAF \citep{Gao2018DynamicFW} & 64.03 & 48.82 & 63.55 & 65.88 & 64.49 & 64.11 & 57.12 & 67.17 & 60.72\\
ViLT \citep{pmlr-v139-kim21k} & 60.48 & 63.89 & 60.27 & 63.20 & 61.38 & 57.00 & 60.72 & 61.90 & 61.14\\
Patch-TRM \citep{lu2022iconqanewbenchmarkabstract} & 65.19 & 46.79 & 65.55 & 66.96 & 55.28 & 64.95 & 58.04 & 67.50 & 61.42 \\
VisualBERT \citep{li2019visualbert} & 59.33 & 69.18 & 61.18 & 62.71 & 62.17 & 58.54 & 62.96 & 59.92 & 61.87 \\ 
UnifiedQA Base \citep{khashabi-etal-2020-unifiedqa} & 68.16 & 69.18 & 74.91 & 63.78 & 61.38 & 77.84 & 72.98 & 65.00 & 70.12 \\\midrule
GPT-4 w/ CoT \citep{openai2024gpt4technicalreport} & 85.48 & 72.44& 90.27 & 82.65 & 71.49 & 92.89 & 86.66 & 79.04 & 83.99\\
GPT4o~\citep{openai2024gpt4technicalreport} & 92.72 & 90.66 & 92.09 & 89.24 & 88.45 & 93.38 & 94.13 & 88.53 & 92.13\\
Chameleon \citep{10.5555/3666122.3668004} & 89.83 & 74.13 & 89.82 & 88.27 & 77.64 & 92.13 & 88.03 & 83.72 & 86.54\\\midrule
LLaMA-Adapter \citep{zhang2024llamaadapterefficientfinetuninglanguage} & 84.37 & 88.30 & 84.36 & 83.72 & 80.32 & 86.90 & 85.83 & 84.05 & 85.19 \\
LaVIN-13B \citep{10.5555/3666122.3667410} & 89.88 & 94.49 & 89.92 & 88.95 & 87.61 & 91.85 & 91.45 & 89.72 & 90.83\\\midrule
Phi-3 Vision~\citep{abdin2024phi3technicalreporthighly} & 75.89 & 85.60 & 49.73 & 85.81 & 80.37 & 50.11 & 75.70 & 62.95 & 71.14 \\
MOLMo 7B-D~\citep{deitke2024molmopixmoopenweights} & 85.66 & 90.33 & 69.00 & 88.09 & 87.95 & 71.22 & 85.24 & 77.06 & 82.32 \\
Qwen2-VL 7B~\citep{yang2024qwen2technicalreport} & 87.97 & 85.49 & 78.73 & 93.41 & 84.29 & 81.53 & 89.72 & 76.67 & 85.05 \\
Pixtral 12B~\citep{agrawal2024pixtral12b} & 87.35 & 92.13 & 79.27 & 92.78 & 87.06 & 81.53 & 88.55 & 82.14 & 86.25 \\\midrule
KAM-CoT \citep{mondal2024kamcotknowledgeaugmentedmultimodal} & 94.76 & 82.24 & \underline{93.36} & 94.53 & 93.16 & \underline{94.15} & 94.24 & 93.21 & 93.87\\
T-SciQ \citep{10.1609/aaai.v38i17.29884} & \underline{96.89} & 95.16 & \textbf{95.55} & \underline{96.53} & 94.70 & \textbf{96.79} & \textbf{96.44} & \underline{95.72} & \textbf{96.18}\\ \midrule
LLaVA OneVision 7B~\citep{li2025llavaonevision} & 88.10 & 88.98 & 78.64 & 92.40 & 87.31 & 80.14 & 88.95 & 80.22 & 85.83 \\
LLaVA-SciGram 7B & 96.27 & \underline{97.53} & 91.64 & \textbf{99.11} & \underline{95.24} & 93.38 & 95.49 & 95.06 & 95.33 \\
LLaVA-SciGram OneVision 7B & \textbf{97.34} & \textbf{99.10} & 90.91 & \textbf{99.11} & \textbf{97.62} & 92.13 & \underline{96.15} & \textbf{95.85} & \underline{96.04} \\
\bottomrule
\end{tabular}
\end{adjustbox}
\caption{Performance comparison on the SQA test set per question type and overall. Values denote accuracy (\%). Bold indicates the best result; underlined the second best.} 
\label{table:sqaresults}
\end{table}

SQA results (Table~\ref{table:sqaresults}) show that our models achieve a new state of the art on visual support questions (IMG), surpassing the previous best by 0.54\% (LLaVA-SciGram) and 2.92\% (LLaVA-SciGram OV). 
LLaVA-SciGram OV reaches 
accuracy comparable to the best model, T-SciQ~\citep{10.1609/aaai.v38i17.29884}, 
specialized for SQA and 
using chain-of-thought reasoning.

\begin{table}[ht]
\small
\centering
\begin{tabular}{lccc}
Model & Opaque Labels & Transparent Labels & Overall \\ \midrule
Phi-3 Vision~\citep{abdin2024phi3technicalreporthighly} & 74.19 & 78.10 & 76.15\\
MOLMo 7B-D~\citep{deitke2024molmopixmoopenweights} & \underline{82.40} & \underline{93.20} & \underline{87.80}\\
Pixtral 12B~\citep{agrawal2024pixtral12b} & 76.46 & 79.00 & 77.73\\
Qwen2-VL 7B~\citep{yang2024qwen2technicalreport} & 80.57 & 83.00 & 81.79 \\ \midrule
Gemini 2.0 Flash~\citep{geminiteam2025geminifamilyhighlycapable} & 73.01 & 83.50 & 78.26\\
GPT4o~\citep{openai2024gpt4technicalreport} & 74.61 & \textbf{94.20} & 84.41\\\midrule
LLaVA OneVision 7B~\citep{li2025llavaonevision} & 79.50 & 89.90 & 84.70\\
LLaVA-SciGram 7B & 80.21 & 89.93 & 85.07\\
LLaVA-SciGram OneVision 7B & \textbf{83.45} & 92.42 & \textbf{87.94}\\
\bottomrule
\end{tabular}
\caption{Performance comparison on the AI2D test set. Values denote accuracy (\%). Bold indicates the best result; underlined indicates the second best.}
\label{table:ai2dresults}
\end{table}

On AI2D (Table~\ref{table:ai2dresults}), LLaVA-SciGram OV surpasses MOLMo, the previous SotA on the opaque-label split, by 1.05\%, showing strong understanding of diagram elements and processes. On the less challenging transparent-label split, GPT4o leads, followed closely by MOLMo and our model. Overall, LLaVA-SciGram OV performs strongest across both splits.

\subsection{Ablation study}

To assess the impact of each SciGram subset within the OV training pipeline, we compare them against their corresponding OV counterparts. As shown in Table~\ref{table:ablationresults1}, using SciGram data consistently improves performance on SQA and AI2D. On TQA DQ, our model surpasses the baseline in the first two stages and remains on par in the final stage, despite using substantially fewer instructions.

\begin{table}[ht]
\centering
\small
\begin{adjustbox}{width=\textwidth}
\begin{tabular}{lllccc}
Alignment & Instruction Tuning & Further Finetuning & TQA DQ & SQA IMG & AI2D Op. \\ \midrule
BLIP Alignment (0.5M inst.) & - & - & 36.86 & 64.80 & 45.43 \\ 
\textcolor{BrickRed}{\textbf{SciGram-Align}} (0.5M inst.) & - & - & \textbf{50.72} & \textbf{72.68} & \textbf{59.07} \\ \midrule
BLIP Alignment & Mid Stage Data (4M inst.) & - & 60.49 & 74.57 & 61.66 \\ 
\textcolor{BrickRed}{\textbf{SciGram-Align}} & \textcolor{ForestGreen}{\textbf{SciGram-VIT}} (0.74M inst.) & - & \textbf{72.06} & \textbf{81.91} & \textbf{74.61} \\ \midrule
BLIP Alignment & Mid Stage Data & OV Single-Img (3.2M inst.) & \textbf{77.08} & 87.31 & 79.50 \\ 
\textcolor{BrickRed}{\textbf{SciGram-Align}} & \textcolor{ForestGreen}{\textbf{SciGram-VIT}} & \textcolor{MidnightBlue}{\textbf{SciGram-M\textsuperscript{3}}} (48K inst.) & 76.68 & \textbf{95.24} & \textbf{80.21} \\
\bottomrule
\end{tabular}%
\end{adjustbox}
\caption{Performance of LLaVA trained on LLaVA OV vs. SciGram across fine-tuning stages. The numbers in parentheses denote the number of instructions contained in each subset.}
\label{table:ablationresults1}
\end{table}

To assess the contribution of each subset, Table~\ref{table:ablationresults2} evaluates
SciGram subset combinations when fine-tuning LLaVA OV 7B. The strongest results use the full pipeline (\textcolor{BrickRed}{\textbf{Align}},  \textcolor{ForestGreen}{\textbf{VIT}}, \textcolor{MidnightBlue}{\textbf{M\textsuperscript{3}}}), showing that all subsets contribute. We also show that adding text-only datasets (OpenBookQA, ARC) to \textcolor{MidnightBlue}{\textbf{SciGram-M\textsuperscript{3}}} provides small but consistent gains across benchmarks.

\begin{table}[ht]
\small
\centering
\begin{tabular}{lllccc}
Alignment & Instruction Tuning & Further Finetuning & TQA DQ & SQA IMG & AI2D Op. \\ \midrule
\textcolor{BrickRed}{\textbf{SciGram-Align}} & - & - & 70.99 & 82.10 & 70.56 \\
- & \textcolor{ForestGreen}{\textbf{SciGram-VIT}} & - & 76.50 & 85.71 & 82.61 \\
- & - & \textcolor{MidnightBlue}{\textbf{SciGram-M\textsuperscript{3}}} & 77.93 & 96.98 & 82.32 \\
\textcolor{BrickRed}{\textbf{SciGram-Align}} & \textcolor{ForestGreen}{\textbf{SciGram-VIT}} & - & 76.65 & 86.47 & 81.83 \\
\textcolor{BrickRed}{\textbf{SciGram-Align}} & - & \textcolor{MidnightBlue}{\textbf{SciGram-M\textsuperscript{3}}} & 77.47 & 94.40 & 81.32 \\
- & \textcolor{ForestGreen}{\textbf{SciGram-VIT}} & \textcolor{MidnightBlue}{\textbf{SciGram-M\textsuperscript{3}}} & 79.94 & 97.47 & 83.42 \\
\textcolor{BrickRed}{\textbf{SciGram-Align}} & \textcolor{ForestGreen}{\textbf{SciGram-VIT}} & \textcolor{MidnightBlue}{\textbf{SciGram-M\textsuperscript{3}}*} & 79.91 & 97.57 & 82.90 \\
\textcolor{BrickRed}{\textbf{SciGram-Align}} & \textcolor{ForestGreen}{\textbf{SciGram-VIT}} & \textcolor{MidnightBlue}{\textbf{SciGram-M\textsuperscript{3}}} & \textbf{80.12} & \textbf{97.62} & \textbf{83.45} \\
\bottomrule
\end{tabular}%
\caption{LLaVA OV results with SciGram subset combinations. *excludes text-only datasets.}
\label{table:ablationresults2}
\end{table}

\subsection{Diagram Comprehension Analysis}
\label{sec:diagram_comprehension}

To better understand how SciGram improves diagram comprehension, we classify a random sample of 200 questions from the TQA DQ test set into eight knowledge types and nine reasoning types, following the ARC taxonomy proposed by~\citet{Clark2018ThinkYH}. We exclude the \textit{Experiments} knowledge type and the \textit{Analogy} reasoning type, as they are not represented in our sample. We also introduce an additional knowledge type, \textit{Visual Cue}, to capture questions that require identifying visual properties such as colors or shapes in the diagram (e.g., "What is the dark blue cell material called?", "How is the name of the star-like organelle inside the large central vacuole?"), and a new reasoning type, \textit{Visual Labeling}, which captures questions where diagram labels are replaced with symbols or letters that must be mapped to the correct entities (e.g., "Which letter represents the ribosome?").

\begin{figure}[ht]
\centering
\includegraphics[width=0.7\linewidth]{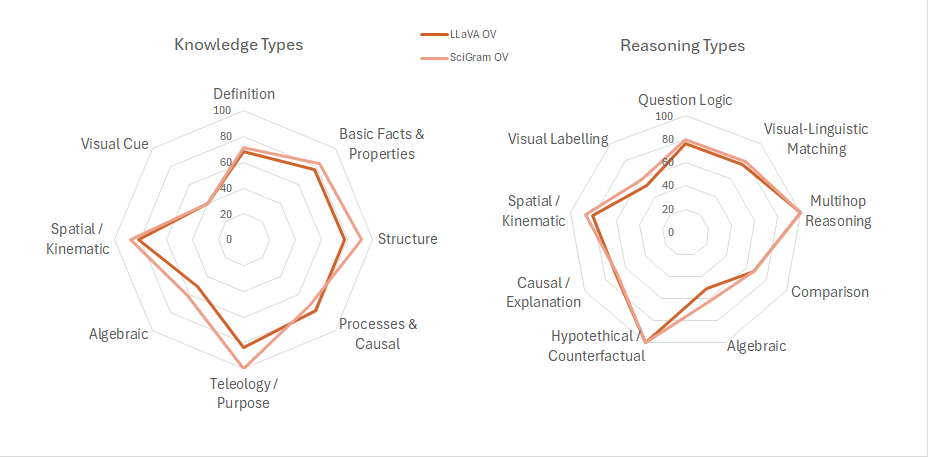}
\caption{LLaVA OV vs LLaVA-SciGram OV by knowledge and reasoning type.} 
\label{fig:knowledge_reasoning_results}
\end{figure}

Figure~\ref{fig:knowledge_reasoning_results} shows that SciGram models match or surpass the baseline across most knowledge and reasoning types, except for \textit{Processes \& Causal} and \textit{Causal/Explanation}. Notably, LLaVA-SciGram OV gains over five points on question types requiring deep diagram understanding, including \textit{Structure}, \textit{Teleology/Purpose}, \textit{Algebraic}, \textit{Spatial/Kinematic}, and \textit{Visual Labeling}. These results demonstrate that SciGram fine-tuning strengthens multiple visual reasoning capabilities aligned with diagram comprehension, while leaving room for further improvement in process and causal-oriented questions.

\subsection{Probing visual grounding}

\begin{wrapfigure}{r}{0.44\textwidth}
\vspace{-30pt}
\centering
\includegraphics[width=\linewidth]{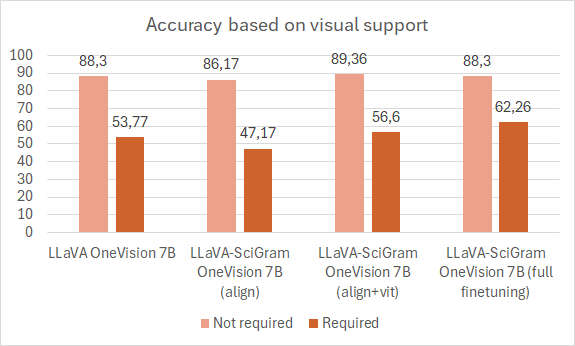}
\caption{LLaVA OV vs LLaVA-SciGram OV (and ablations) on questions requiring and not requiring visual support to be answered.}
\label{fig:visual_support_results}
\vspace{-20pt}
\end{wrapfigure}

A central concern in multimodal QA is whether models genuinely leverage visual inputs or rely on language priors. Using the same 200 TQA questions from Section~\ref{sec:diagram_comprehension}, we annotate whether each requires visual support to be correctly answered. For instance, a question such as "How many phases of meiosis I are shown in the diagram?" requires analyzing the image, whereas a question such as "What do you call the group of protozoans characterized by the presence of hair-like organelles called cilia?" can be answered through prior textual knowledge. From them, 94 questions were labeled \textit{visual support not required}.

Figure~\ref{fig:visual_support_results} shows that LLaVA-SciGram OV outperforms LLaVA OV by nearly ten points on questions requiring visual reasoning, while matching it on questions that can be answered from language priors. Progressive fine-tuning across SciGram subsets further improves performance, with the full pipeline yielding the strongest results. While further qualitative evaluations, including image ablations and diagram swapping, are left for future work, these results suggest that SciGram's gains stem from improved diagram understanding rather than purely textual cues.

\section{Conclusions and future work}
\label{sec:conc}

In this paper, we present a framework for generating large-scale visual instruction data for diagrams that leverages curriculum-derived scientific terminology. Using this framework, we created SciGram, a dataset containing over 194k diagrams and 1.4M visual instructions in the natural sciences. Our experiments demonstrate that models fine-tuned on SciGram achieve substantial improvements on diagram-centric benchmarks such as TQA, SQA, and AI2D, outperforming or matching existing state-of-the-art vision–language models while using substantially fewer training instances. Moreover, we show that further training of existing models like LLaVA OneVision with SciGram can establish new state-of-the-art performance in diagram-based question answering. This illustrates that, despite some noise and minor inconsistencies, SciGram provides a strong signal for learning visual instructions without costly manual curation. Beyond benchmark performance, our results also suggest that SciGram improves visual grounding and diagram-centric reasoning rather than merely 
textual knowledge, supporting its effectiveness as a source of multimodal supervision.

Future work will focus on improving dataset quality through more precise diagram filtering, better caption–diagram alignment, improved question generation, and enhanced factual accuracy of synthesized scientific claims. Additionally, our model-agnostic methodology can naturally benefit from stronger teacher models as they become available, enabling continued improvements in dataset quality. More broadly, we hope that the SciGram methodology will provide a practical foundation for developing future domain-specialized vision-language models beyond scientific diagram understanding. While demonstrated here for scientific diagrams, the same methodology could be extended to other structured visual knowledge sources, such as engineering schematics, medical illustrations, or educational graphics, offering a scalable approach for constructing domain-specific multimodal supervision without relying on costly manual annotation.

\clearpage

\section*{Ethics Statement}
\label{sec:ethics}

\textbf{Licensing and Data Usage.}
All datasets and pretrained models are subject to their respective licenses, and future users are responsible for complying with their terms. Improper use of copyrighted datasets or proprietary models may result in legal or ethical violations. As noted in our GitHub repository on the license and copyright of content linked from SciGram:
\begin{itemize}[leftmargin=*]
\item Images linked from SciGram are copyrighted by their respective owners; the SciGram authors do not host or redistribute them.
\item Image URLs are publicly available on the internet and were not scraped from private sources.
\item We respected robots.txt rules and site Terms of Service (ToS) during URL collection.
\item SciGram is intended for educational and research purposes only; its creators do not claim ownership of linked content.
\item SciGram released under CC BY 4.0 (\url{https://creativecommons.org/licenses/by/4.0}).
\end{itemize}

\textbf{Bias and Fairness.}
Pretrained models may reflect biases in their training data. Although our study focuses on diagram reasoning, such biases may affect downstream outputs, potentially disadvantaging certain groups or misrepresenting information. Users should consider these risks when deploying similar models.

\textbf{Environmental Impact.}
Training and fine-tuning large models are computationally expensive and contribute to carbon emissions. We encourage efficient training strategies and consideration of environmental costs when developing similar systems.

\textbf{Misuse Potential.}
Although intended for research and educational purposes, our approach could be misused for automated content generation or misinformation. Appropriate safeguards and ethical guidelines should be followed to minimize potential harm.

\section*{Reproducibility Statement}
\label{sec:reproducibility}

We release the code used to generate SciGram, train the models, and replicate the evaluations: \url{https://github.com/expertailab/scigram}. Data and models are available at \url{https://huggingface.co/collections/expertailab/scigram}. Nevertheless, several factors may limit full reproducibility:

\textbf{Link rot and variable availability.}
Due to licensing and copyright restrictions, we distribute only the URLs of images in SciGram. Consequently, some images may become inaccessible over time, as required by the terms of use of their original sources.

\textbf{Hardware constraints.}
Our framework relies on large pre-trained models requiring substantial resources for fine-tuning and inference, which may limit accessibility for researchers with limited hardware. Scaling the architectures or models used to generate and train SciGram would require even greater resources.

\textbf{Use of external APIs.}
Some evaluation metrics rely on proprietary APIs that may incur costs and change or be discontinued over time, making exact reproduction of evaluation results challenging and limiting long-term comparability.


\section*{Acknowledgments}
This work was supported by the Digital Europe Programme through LLMs4EU (Grant Agreement No. 101198470) and the Horizon Europe project FAIR2Adapt (Grant Agreement No. 101188256). GPU infrastructure was provided by IPCEI-CIS – Progetto Villanova (Prog. n. SA.102519 – CUP B29J24000850005) and INESData (Infrastructure to Investigate Data Spaces in Distributed Environments at UPM), funded under the UNICO I+D CLOUD call by the Ministry for Digital Transformation and the Civil Service within the PRTR recovery plan financed by the European Union (NextGenerationEU). We also thank Flavio Merenda for valuable feedback on successive manuscript versions.

\clearpage
\bibliography{colm2026_conference}
\bibliographystyle{colm2026_conference}

\clearpage
\appendix
\section{SciGram Examples}
\label{sec:complete_examples}
\begin{figure*}[ht]
\centering

\includegraphics[width=\linewidth]{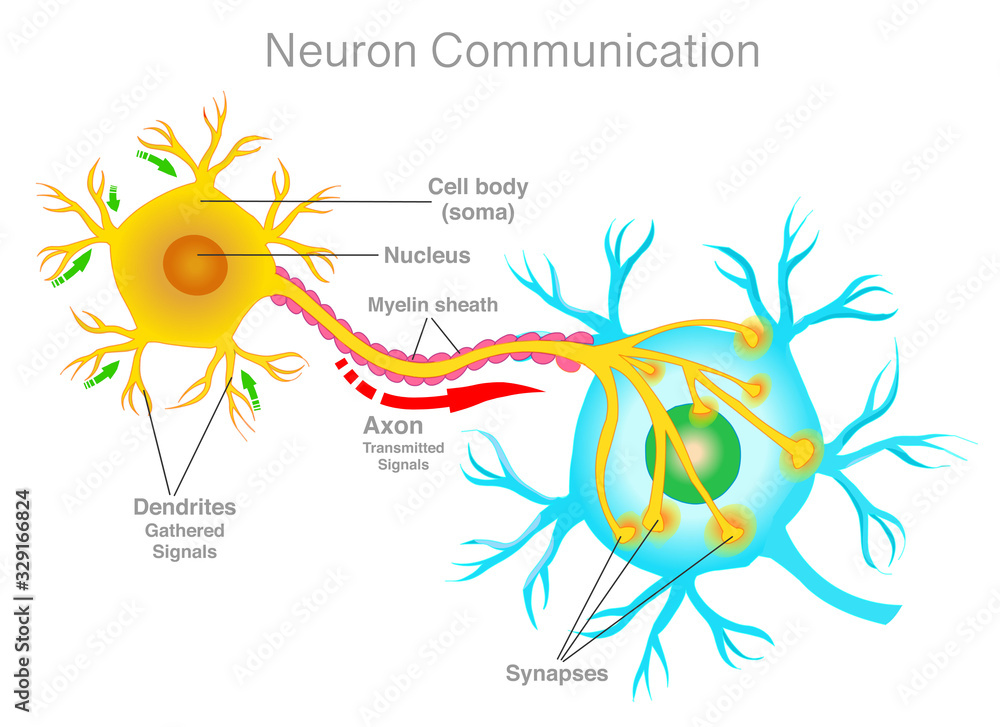}

\noindent
\begin{minipage}[t]{0.54\linewidth}
\vspace{0pt}
\tcbset{
    breakable,
    colback=BrickRed!5!white,
    colframe=BrickRed!75!black,
    boxrule=0.8pt,
    arc=4pt,
    left=2mm,
    right=2mm,
    top=1mm,
    bottom=1mm
}
\begin{tcolorbox}[title=SciGram-Align (captions)]
\tiny

\textbf{C1.} The diagram illustrates the process of neuron communication, showcasing the anatomy of two neurons and the flow of signals between them. On the left, a neuron with a yellow cell body (soma) and a green nucleus is depicted. This neuron has dendrites extending from its cell body, which are responsible for gathering signals from other neurons. The signals are then transmitted through the axon, a long, thin projection that extends from the cell body. The axon is covered by a myelin sheath, which helps to insulate the axon and speed up the transmission of signals. The axon terminates at the synapses, which are the points of contact between neurons. On the right, another neuron with a blue cell body (soma) and a green nucleus is shown. The signals from the first neuron are transmitted across the synapses to the second neuron, which then processes the signals and passes them on to other neurons. The diagram highlights the importance of the synapses in facilitating the communication between neurons and underscores the intricate network of connections that underlie neural function.

\textbf{C2.} The diagram illustrates the process of neuron communication within the nervous system. It features two neurons, one with a yellow soma and the other with a blue soma, connected by an axon. The yellow neuron's dendrites, which are the signal-gathering parts, are depicted with green arrows pointing towards the soma. These dendrites are responsible for receiving signals from other neurons. The axon, which is the signal-transmitting part, is shown extending from the soma of the yellow neuron and connecting to the blue neuron's dendrites. The axon is surrounded by a myelin sheath, which is a protective layer that speeds up the transmission of signals. The signals are transmitted from the yellow neuron to the blue neuron through synapses, which are the points of contact between neurons. The blue neuron's dendrites, which are the receiving parts, are shown with yellow arrows pointing towards the soma. The nucleus of each neuron is also labeled, indicating the control center of the cell.

\textbf{C3.} The diagram illustrates the process of neuron communication within the nervous system. It shows two neurons connected by an axon, which is the long, thin projection that transmits signals from the cell body to other neurons. The cell body, also known as the soma, contains the nucleus and is the central part of the neuron. Dendrites, which are the tree-like structures on the left neuron, gather signals from other neurons and transmit them to the cell body. The myelin sheath, a fatty coating around the axon, helps to speed up the transmission of signals. The axon then extends to the right neuron, where it synapses with the dendrites of the target neuron. The transmitted signals are indicated by the red arrow, showing the direction of signal propagation from one neuron to the next.

\end{tcolorbox}
\end{minipage}
\hfill
\begin{minipage}[t]{0.43\linewidth}
\vspace{0pt}
\tcbset{
    colback=ForestGreen!5!white,
    colframe=ForestGreen!75!black
}
\begin{tcolorbox}[title=SciGram-VIT (MCQs)]
\scriptsize

\textbf{Q1. What happens at the synapses?}\\
a) The signal is received \\
b) The signal is received and transmitted \\
c) The signal is generated \\
d) The signal is transmitted $\checkmark$

\textbf{Q2. What is the role of the myelin sheath?}\\
a) Protects the cell body \\
b) Carries signals from the cell body to the axon \\
c) Carries signals from dendrites to the cell body \\
d) Insulates the axon and speeds transmission $\checkmark$

\textbf{Q3. Which part transmits signals between neurons?}\\
a) Myelin sheath \\
b) Cell body \\
c) Dendrites \\
d) Synapses $\checkmark$

\textbf{Q4. Where do signals from dendrites enter?}\\
a) Myelin sheath \\
b) Cell body $\checkmark$ \\
c) Axon \\
d) Nucleus

\textbf{Q5. What is the function of dendrites?}\\
a) Transmit signals \\
b) Receive signals $\checkmark$ \\
c) Store signals \\
d) Generate signals

\end{tcolorbox}
\end{minipage}

\caption{
Example \#1 of SciGram-Align and SciGram-VIT
}

\label{fig:scigram_example1}
\end{figure*}

\begin{figure*}[ht]
\centering

\includegraphics[width=\linewidth]{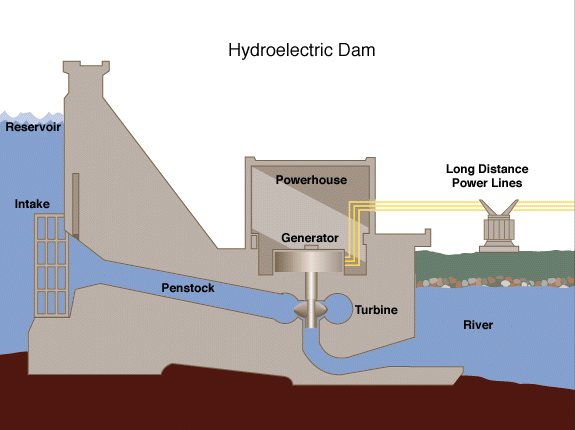}

\noindent
\begin{minipage}[t]{0.55\linewidth}
\vspace{0pt}
\tcbset{
    breakable,
    colback=BrickRed!5!white,
    colframe=BrickRed!75!black,
    boxrule=0.8pt,
    arc=4pt,
    left=2mm,
    right=2mm,
    top=1mm,
    bottom=1mm
}
\begin{tcolorbox}[title=SciGram-Align (captions)]
\scriptsize

\textbf{C1.} The diagram depicts a hydroelectric dam, illustrating its main components and the flow of water through the system. At the top, there is a reservoir, which stores water. The water is then directed through an intake structure, which channels it into the powerhouse. Inside the powerhouse, the water flows into a turbine, which is connected to a generator. The generator converts the kinetic energy of the water into electrical energy. The generated electricity is then transmitted through long-distance power lines to a location where it can be used. The diagram also shows the river, which provides the water for the dam.

\textbf{C2.} The diagram depicts a cross-sectional view of a hydroelectric dam, illustrating the various components and their interactions. At the top, there is a reservoir, which is connected to the intake, allowing water to flow into the dam. The water then passes through a penstock, a large pipe that directs the water to the powerhouse. Inside the powerhouse, there is a turbine, which is connected to a generator. The turbine converts the kinetic energy of the water into mechanical energy, which is then used to rotate the generator. The generator produces electricity, which is transmitted through long distance power lines to the consumer. The powerhouse is situated between the penstock and the turbine, and the generator is located within the powerhouse. The river is shown at the bottom of the diagram, providing the water source for the dam.

\textbf{C3.} The diagram depicts a hydroelectric dam with its various components and their functions. At the top, the reservoir is shown, which holds water that is then directed through an intake. The water then flows through a penstock, a large pipe that channels the water to the powerhouse. Inside the powerhouse, a turbine is situated, which converts the kinetic energy of the water into mechanical energy. The turbine is connected to a generator, which produces electrical energy. The generated electricity is then transmitted through long distance power lines to the consumer.

\end{tcolorbox}
\end{minipage}
\hfill
\begin{minipage}[t]{0.42\linewidth}
\vspace{0pt}
\tcbset{
    colback=ForestGreen!5!white,
    colframe=ForestGreen!75!black
}
\begin{tcolorbox}[title=SciGram-VIT (MCQs)]
\scriptsize

\textbf{Q1. What happens to the water after it passes through the turbine?} \\
a) It is used for irrigation\\
b) It is released into the river $\checkmark$\\
c) It is stored in the reservoir\\
d) It is heated for geothermal energy

\textbf{Q2. What is the role of the generator in a hydroelectric dam?} \\
a) Generates heat for heating purposes\\
b) Catches fish\\
c) Stores water for future use\\
d) Converts mechanical energy into electrical energy $\checkmark$

\textbf{Q3. Where does the water from the reservoir flow through before reaching the turbine?} \\
a) Powerhouse \\
b) Penstock $\checkmark$ \\
c) Long Distance Power Lines \\
d) Intake

\textbf{Q4. What is the function of the penstock in a hydroelectric dam?}  \\
a) Carries fish to the river\\
b) Carries sediment to the river\\
c) Carries electricity to the powerhouse\\
d) Carries water to the turbine $\checkmark$

\textbf{Q5. What is the purpose of the powerhouse in a hydroelectric dam?}\\
a) Provides a habitat for aquatic life\\
b) Generates electricity $\checkmark$\\
c) Catches fish\\
d) Stores water for irrigation

\end{tcolorbox}
\end{minipage}

\caption{
Example \#2 of SciGram-Align and SciGram-VIT
}

\label{fig:scigram_example2}
\end{figure*}

\begin{figure*}[ht]
\centering

\includegraphics[width=\linewidth]{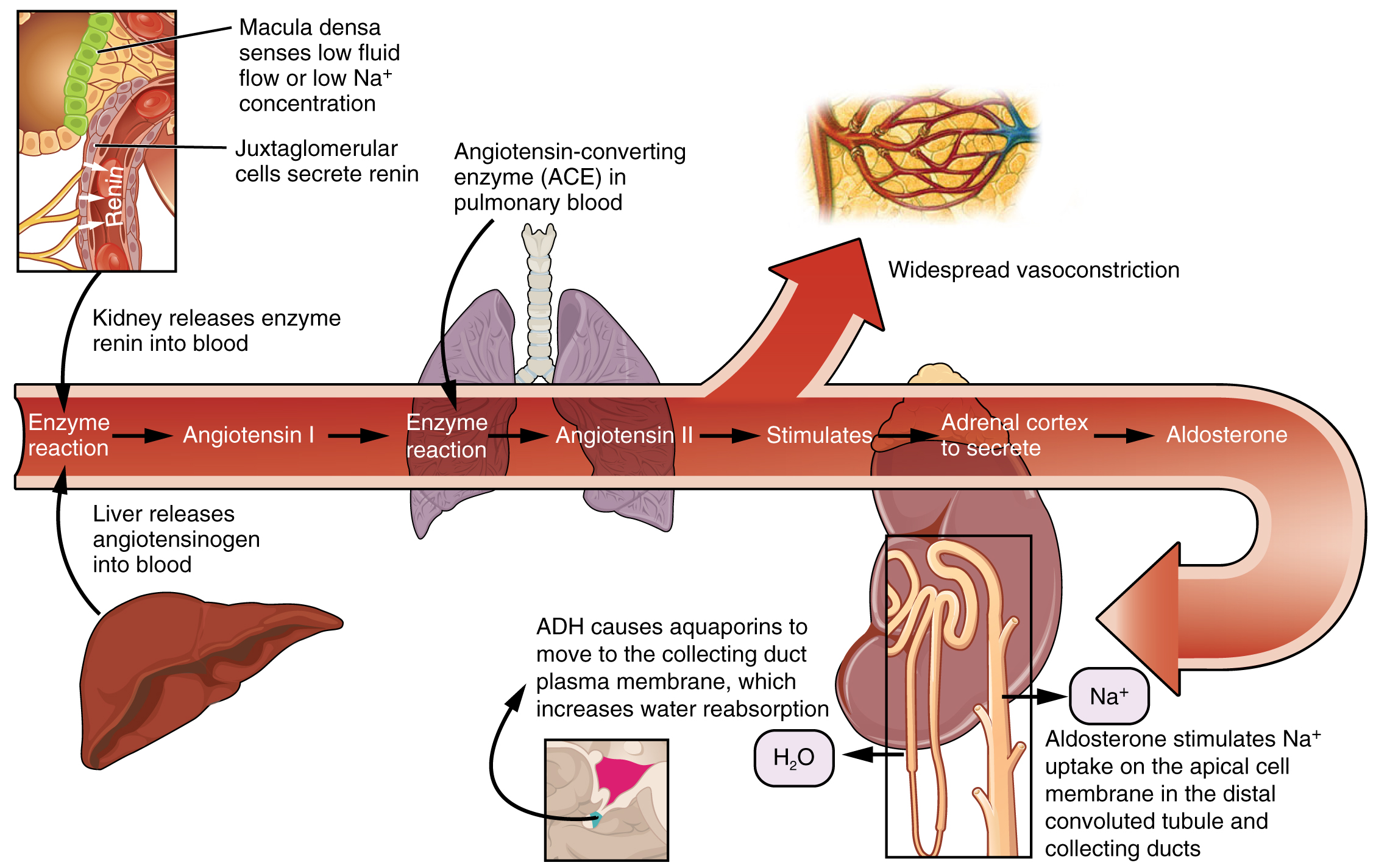}

\noindent
\begin{minipage}[t]{0.55\linewidth}
\vspace{0pt}
\tcbset{
    colback=BrickRed!5!white,
    colframe=BrickRed!75!black,
    boxrule=0.8pt,
    arc=4pt,
    left=2mm,
    right=2mm,
    top=1mm,
    bottom=1mm
}
\begin{tcolorbox}[title=SciGram-Align (captions)]
\scriptsize

\textbf{C1.} The diagram explains the renin-angiotensin-aldosterone system. The process begins with the kidney releasing the enzyme renin into the blood. The liver releases angiotensinogen into the blood. The angiotensinogen is converted to angiotensin I by an enzyme reaction. Angiotensin-converting enzyme (ACE) in the pulmonary blood converts angiotensin I to angiotensin II. Angiotensin II stimulates the adrenal cortex to secrete aldosterone. Aldosterone stimulates sodium uptake on the apical cell membrane in the distal convoluted tubule and collecting ducts. This leads to widespread vasoconstriction. ADH causes aquaporins to move to the collecting duct plasma membrane, which increases water reabsorption.

\textbf{C2.} The diagram shows the process of aldosterone secretion in the body. The kidney releases enzyme renin into the blood, which then travels through the bloodstream to the liver, where it is converted to angiotensin I. The angiotensin I then travels to the lungs, where it is converted to angiotensin II. Angiotensin II stimulates the adrenal cortex to secrete aldosterone. Aldosterone then travels to the kidneys, where it stimulates the uptake of sodium ions (Na+) on the apical cell membrane in the distal convoluted tubule and collecting ducts. This process leads to increased water reabsorption, which helps to maintain fluid balance in the body.

\textbf{C3.} The diagram shows the renin-angiotensin-aldosterone system, which is a hormonal system that regulates blood pressure and fluid balance in the body. The process begins when the macula densa in the kidney senses low fluid flow or low sodium concentration, triggering the release of renin by the juxtaglomerular cells. Renin then converts angiotensinogen, released by the liver, into angiotensin I. This is then converted to angiotensin II by the angiotensin-converting enzyme (ACE) in the pulmonary blood. Angiotensin II stimulates the adrenal cortex to secrete aldosterone, which in turn increases sodium uptake on the apical cell membrane in the distal convoluted tubule and collecting ducts. This leads to increased reabsorption of sodium and water, resulting in vasoconstriction and increased blood pressure. The diagram also shows that ADH causes aquaporins to move to the collecting duct plasma membrane, which increases water reabsorption.

\end{tcolorbox}
\end{minipage}
\hfill
\begin{minipage}[t]{0.43\linewidth}
\vspace{0pt}
\tcbset{
    colback=ForestGreen!5!white,
    colframe=ForestGreen!75!black
}
\begin{tcolorbox}[title=SciGram-VIT (MCQs)]

\scriptsize

\textbf{Q1. Where does angiotensin-converting enzyme (ACE) play a role in the renin-angiotensin-aldosterone system?} \\
a) Kidney \\
b) Liver \\
c) Pulmonary blood $\checkmark$\\
d) Adrenal cortex

\textbf{Q2. Which hormone causes aquaporins to move to the collecting duct plasma membrane?} \\
a) Antidiuretic hormone (ADH) $\checkmark$ \\
b) Renin \\
c) Aldosterone \\
d) Angiotensin II

\textbf{Q3. What is the effect of aldosterone on sodium (Na+) uptake in the distal convoluted tubule and collecting ducts?} \\
a) Does not affect Na+ uptake \\
b) Stimulates Na+ uptake $\checkmark$ \\
c) Causes Na+ to be excreted \\
d) Decreases Na+ uptake

\textbf{Q4. What happens when the kidney releases renin into the blood?} \\
a) Angiotensin I is converted to angiotensin II $\checkmark$ \\
b) Aldosterone is secreted by the adrenal cortex \\
c) Water reabsorption decreases \\
d) Widespread vasoconstriction

\textbf{Q5. What is the role of the Macula densa in the renin-angiotensin-aldosterone system?} \\
a) Senses low fluid flow or low Na+ concentration $\checkmark$ \\
b) Stimulates the adrenal cortex to secrete aldosterone \\
c) Secretes renin \\
d) Converts angiotensin I to angiotensin II

\end{tcolorbox}
\end{minipage}

\caption{
Example \#3 of SciGram-Align and SciGram-VIT
}

\label{fig:scigram_example3}
\end{figure*}

Figures~\ref{fig:scigram_example1}-\ref{fig:scigram_example3} show examples of diagrams from the SciGram dataset, along with all the captions and MCQs generated using our methodology.

\tcbset{reset}

\section{Prompts and model configurations}
\label{sec:prompts}

\subsection{Atomic fact generation}

\tcbset{
    breakable,
    colback=gray!5!white,
    colframe=gray!75!black,
    boxrule=0.8pt,
    arc=4pt,
    left=2mm,
    right=2mm,
    top=1mm,
    bottom=1mm
}

\begin{tcolorbox}[title=Prompt template for atomic fact generation]
\begin{lstlisting}[breaklines=true, columns=flexible, basicstyle=\small, upquote=false]
<|begin_of_text|><|start_header_id|>system<|end_header_id|>
Your task is to generate a list of 50 atomic facts that contain different combinations of the given terms.

Please follow this criteria:
- Limit yourself to the scientific domains of life sciences, earth sciences, and physical sciences at the middle school level.
- Pay attention to commonsense.
- Make the facts brief and concise yet easy to understand and meaningful.
- Use different grammatical constructions and limit length to 12 tokens or less.

Arrange your output as a jsonl file where each line is {{"atomic fact": <fact>}}.

<|begin_of_text|><|start_header_id|>user<|end_header_id|>
TERMS: ["atom", "proton", "neutron"].

OUTPUT: {{"output": "<INSERT OUTPUT HERE>"}}
\end{lstlisting}
\end{tcolorbox}

\begin{configbox}
Temperature   : 0.7 \\
P value       : 1.0 \\
Max length    : 1024
\end{configbox}

\subsection{Caption Generation}

\tcbset{
    breakable,
    colback=gray!5!white,
    colframe=gray!75!black,
    boxrule=0.8pt,
    arc=4pt,
    left=2mm,
    right=2mm,
    top=1mm,
    bottom=1mm
}

\begin{tcolorbox}[title=Prompt template for caption generation]
\begin{lstlisting}[breaklines=true, columns=flexible, basicstyle=\small]
<|im_start|>user
<|vision_start|><|image_pad|><|vision_end|> Provide a paragraph with a brief description of the diagram. Pay special attention to the main components of the diagram and the relations between them. If visible, also reflect space and temporal information, linking it to the components and relations in the diagram.<|im_end|>
<|im_start|>assistant

\end{lstlisting}
\end{tcolorbox}

\begin{configbox}
Temperature   : 0.7 \\
P value       : 0.7 \\
Max length    : 512
\end{configbox}

\subsection{Multiple Choice Question Generation}

\tcbset{
    breakable,
    colback=gray!5!white,
    colframe=gray!75!black,
    boxrule=0.8pt,
    arc=4pt,
    left=2mm,
    right=2mm,
    top=1mm,
    bottom=1mm
}

\begin{tcolorbox}[title=Prompt template for multiple-choice question generation]
\begin{lstlisting}[breaklines=true, columns=flexible, basicstyle=\small, upquote=false]
<|im_start|>user
<|vision_start|><|image_pad|><|vision_end|>
Formulate five multiple choice questions with 4 possible answers grounded in the diagram. The resulting questions must be middle school level questions in the subjects of life sciences, earth sciences or physical sciences. The questions should be answered using the elements from the image. For your output, follow this structure:
[{"question": <question>, "answers": {"a": <answer a>, "b": <answer b>, "c": <answer c>, "d": < answer d>}, "correct_answer": <correct_letter>}, ...].<|im_end|>
<|im_start|>assistant
\end{lstlisting}
\end{tcolorbox}

\begin{configbox}
Temperature   : 0.0 \\
P value       : 0.0 \\
Max length    : 1024
\end{configbox}

\subsection{Question Answering}
\label{sec:qa_prompts}

\tcbset{
    breakable,
    colback=gray!5!white,
    colframe=gray!75!black,
    boxrule=0.8pt,
    arc=4pt,
    left=2mm,
    right=2mm,
    top=1mm,
    bottom=1mm
}

\begin{tcolorbox}[title=Prompt template for multiple-choice question answering with diagrams]
\begin{lstlisting}[breaklines=true, columns=flexible, basicstyle=\small]
<|im_start|>user
<|vision_start|><|image_pad|><|vision_end|>
Take a look at the diagram and answer the following question by choosing one of the possible answers.
Question: What does igneous rock become when it is subjected to heat and pressure?
Answer choices: 
a) Magma
b) Sediments
c) Metamorphic rock
d) Sedimentary rock
<|im_end|>
<|im_start|>assistant
\end{lstlisting}
\end{tcolorbox}

\begin{configbox}
Temperature   : 0.0 \\
P value       : 0.0 \\
Max length    : 4096
\end{configbox}

\begin{tcolorbox}[title=Prompt template for multiple-choice question answering only text without context paragraph]
\begin{lstlisting}[breaklines=true, columns=flexible, basicstyle=\small]
<|im_start|>user
Answer the following question by choosing one of the possible answers.
Question: What does igneous rock become when it is subjected to heat and pressure?
Answer choices: 
a) Magma
b) Sediments
c) Metamorphic rock
d) Sedimentary rock
<|im_end|>
<|im_start|>assistant
\end{lstlisting}
\end{tcolorbox}

\begin{configbox}
Temperature   : 0.0 \\
P value       : 0.0 \\
Max length    : 4096
\end{configbox}

\begin{tcolorbox}[title=Prompt template for multiple-choice question answering only text with context paragraph]
\begin{lstlisting}[breaklines=true, columns=flexible, basicstyle=\small]
<|im_start|>user
Lesson: The volcanoes (...)
Take a look at the lesson and answer the following question by choosing one of the possible answers.
Question: What does igneous rock become when it is subjected to heat and pressure?
Answer choices: 
a) Magma
b) Sediments
c) Metamorphic rock
d) Sedimentary rock
<|im_end|>
<|im_start|>assistant
\end{lstlisting}
\end{tcolorbox}

\begin{configbox}
Temperature   : 0.0 \\
P value       : 0.0 \\
Max length    : 4096
\end{configbox}

\begin{tcolorbox}[title=Prompt template for multiple-choice question answering with diagrams and context paragraph]
\begin{lstlisting}[breaklines=true, columns=flexible, basicstyle=\small]
<|im_start|>user
<|vision_start|><|image_pad|><|vision_end|>
Lesson: The volcanoes (...)
Take a look at the lesson and the diagram, and answer the following question by choosing one of the possible answers.
Question: What does igneous rock become when it is subjected to heat and pressure?
Answer choices: 
a) Magma
b) Sediments
c) Metamorphic rock
d) Sedimentary rock
<|im_end|>
<|im_start|>assistant
\end{lstlisting}
\end{tcolorbox}

\begin{configbox}
Temperature   : 0.0 \\
P value       : 0.0 \\
Max length    : 4096
\end{configbox}

\section{Instruction-Following Examples}
\label{sec:instruction_examples}

\subsection{SciGram-Align instructions}

\tcbset{
    breakable,
    colback=BrickRed!5!white,
    colframe=BrickRed!75!black,
    boxrule=0.8pt,
    arc=4pt,
    left=2mm,
    right=2mm,
    top=1mm,
    bottom=1mm
}

\begin{tcolorbox}[title=Example 1]
\textbf{Human:} I need a detailed caption for this diagram—please include explanations of each section and their functions.\\
\textbf{Assistant:} The diagram consists of two main parts: a textual explanation and a visual representation of an ionic compound. The textual explanation provides a definition of ionic compounds as giant lattice structures of ions held together by strong electrostatic forces. Sodium chloride is used as an example. The visual shows a 3D lattice model with green spheres as chloride ions and purple as sodium ions, arranged in a cubic pattern.
\end{tcolorbox}

\begin{tcolorbox}[title=Example 2]
\textbf{Human:} Please give a thorough caption for this diagram, covering all the important aspects and their significance.\\
\textbf{Assistant:} The diagram explains how cold dark matter forms galaxies, starting from the Big Bang, tracing dark matter halo collapse, gas cloud collapse, proto-spiral formation, and eventual spiral galaxies. Numerical simulations suggest galaxy growth via merging halos filled with cold, non-interacting dark matter.
\end{tcolorbox}

\begin{tcolorbox}[title=Example 3]
\textbf{Human:} Could you elaborate on this diagram with a caption that explains the concepts and relationships shown?\\
\textbf{Assistant:} The diagram shows sand dune formation: embryo dune → fore dune → yellow dune → grey dune. It includes dune slack (vegetation area) and the water table. Progression from sea to land illustrates dune maturation and habitat development.
\end{tcolorbox}

\begin{tcolorbox}[title=Example 4]
\textbf{Human:} Can you create a caption that not only labels but also explains the purpose of this diagram?\\
\textbf{Assistant:} The diagram illustrates the transition between states of matter: solid (dense spheres), liquid (suspended spheres in a beaker), and gas (freely moving spheres in a flask). Each stage shows differences in particle arrangement and movement.
\end{tcolorbox}

\subsection{SciGram-VIT instructions}

\tcbset{
    colback=ForestGreen!5!white,
    colframe=ForestGreen!75!black
}

\begin{tcolorbox}[title=Example 1]
\textbf{Human:} What is the first visible sign of the embryo's development in the diagram?\\
Answer choices:

a) Facial features 

b) Visible elbows 

c) Finger formation 

d) Elongated arms and legs\\
\textbf{Assistant:} \{"answer": "c"\}
\end{tcolorbox}

\begin{tcolorbox}[title=Example 2]
\textbf{Human:} What is the example of a contact force?\\
Answer choices: 

a) Gravitational Force

b) Magnetic Force

c) Spring Force

d) Friction Force\\
\textbf{Assistant:} \{"answer": "d"\}
\end{tcolorbox}

\begin{tcolorbox}[title=Example 3]
\textbf{Human:} What is the name of the part of the trap that is located at the top of the U-shape?\\
Answer choices: 

a) inlet

b) trap weir

c) trap dip

d) trap arm\\
\textbf{Assistant:} \{"answer": "a"\}
\end{tcolorbox}

\begin{tcolorbox}[title=Example 4]
\textbf{Human:} Which planet is closest to the sun?\\
Answer choices: 

a) Mars

b) Mercury

c) Earth

d) Venus\\
\textbf{Assistant:} \{"answer": "b"\}
\end{tcolorbox}

\subsection{SciGram-M3 instructions}

\tcbset{
    colback=MidnightBlue!5!white,
    colframe=MidnightBlue!75!black
}

\begin{tcolorbox}[title=Example]
\textbf{Human:} Take a look at the diagram and answer the following question by choosing one of the possible answers.\\
Question: What does igneous rock become when it is subjected to heat and pressure?\\
Answer choices: 

a) Magma

b) Sediments

c) Metamorphic rock

d) Sedimentary rock\\
\textbf{Assistant:} \{"answer": "c"\}
\end{tcolorbox}

\section{Terminology stats}
\label{sec:terminology_stats}

In this appendix we analyze the scientific terminology used to construct the SciGram dataset. 

Table \ref{table:termsfreq} presents the most and least frequent terms from our final set of 4,820 extracted terms, based on their occurrence in the TQA textbook. Among these, 1,295 terms appear only once, reflecting a long-tail distribution. This skew is mitigated by the weirdness index filter, which retains contextually important terms even if they are infrequent in the source text. In total, 15.09\% of candidate noun phrases were eliminated by this filter.  

Table \ref{table:termswi} shows terms with the highest and lowest weirdness-index scores; terms with infinite weirdness-index (appearing in TQA but absent in the BNC corpus) are not included. Distinctive scientific terms such as “Cellular Respiration” and “Epicenter” score highly, as expected.

\begin{table}
\centering
\small
\begin{tabular}{lc|lc}
\toprule
Term & Frequency & Term & Frequency \\ \midrule
1. Water & 1,930 & (other 1291 terms)  & 1 \\
2. Energy & 1,712 & 4817. Gas Increase & 1 \\
3. Air & 770 & 4818. Temperature and Volume & 1\\
4. Earth & 630 & 4819. Total Spread & 1\\
5. Body & 586 & 4820. Typical Measurement & 1\\
\bottomrule
\end{tabular}%
\caption{Most/least frequent terms in our terminology.}
\label{table:termsfreq}
\end{table}

\begin{table}
\centering
\small
\begin{tabular}{lc|lc}
\toprule
Term & w-index & Term & w-index \\ \midrule
1. Mechanical Advantage & 20,380.0 & 4816. Invention & 2.03 \\
2. Cellular Respiration & 18,391.7 & 4817. Shaft & 2.02 \\
3. Air Mass & 5,799.2 & 4818. Valley & 2.01 \\
4. Epicenter & 5,560.1 & 4819. Heavy Metal & 2.01 \\
5. Decomposers & 4,604.5 & 4820. Rubbing & 2.01 \\
\bottomrule
\end{tabular}
\caption{Terms with the highest and lowest weirdness index in our terminology selection. Infinite weirdness index terms were excluded.}
\label{table:termswi}
\end{table}

\begin{figure}
\centering
\includegraphics[width=0.6\linewidth]{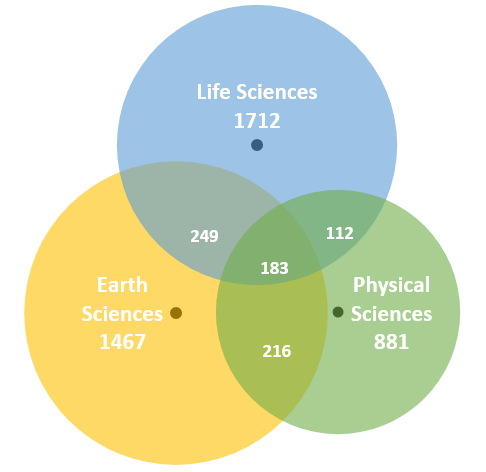}
\caption{Terminology distribution by subjects. For each subject we include the number of terms that appear in lessons from that specific subject. In the intersections, we report the number of terms belonging to two or three matters at the same time. The total number is 4,820.}
\label{fig:termdist}
\end{figure}

Figure \ref{fig:termdist} shows the subject-wise distribution of terminology across the TQA textbook. While Physical Sciences are slightly underrepresented compared to Earth and Life Sciences, there is substantial overlap between subjects: 399 terms are shared between Physical and Earth Sciences, and 432 between Earth and Life Sciences. In general, most terms are assigned to a single subject, but nearly 16\% appear in two subjects, and 183 terms are shared across all three, highlighting both the specificity and the transversal nature of scientific terminology within the middle-school curricula.

\section{Balancing Datasets}
\label{sec:balancing}
\begin{figure}
\centering
\includegraphics[width=0.6\linewidth]{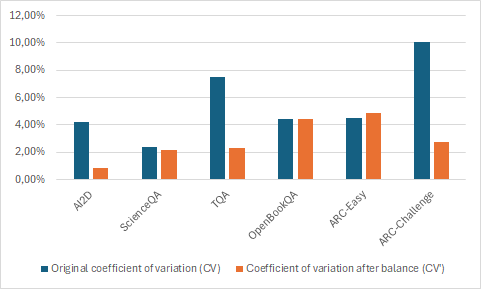}
\caption{Coefficient of variation of correct answers in the SciGram-M\textsuperscript{3} subsets before (CV) and after (CV') balancing.}
\label{fig:cv_scigram}
\end{figure}

Figure~\ref{fig:cv_scigram} shows how the distribution of correct answers is balanced in the SciGram-M\textsuperscript{3} subsets. Originally, datasets such as AI2D or ARC-Challenge present a moderately high coefficient of variation of correct answers within the different answer choices. This can be a source of biases and overfitting while training a model using these datasets, but can be easily alleviated by shuffling these answer choices across the datasets, as showed in the final distribution.

\section{Human evaluation of SciGram}
\label{sec:human_eval}

In this appendix, we present the detailed quality assessment of SciGram conducted by four NLP experts. We aimed to evaluate the quality of the diagrams, captions and multiple-choice questions which are part of SciGram. Thus, we prepared a questionnaire that we think can define the quality of the mentioned elements from the dataset. Each annotator is provided with 200 diagrams, 200 captions-image pairs, and 200 multiple-choice questions with diagrams. We collected the results of the annotators and calculate the average results for each question. The questions and their corresponding results are presented below. All the annotations have a \textit{p}-value $<$ 0.0001.

\subsection{Diagram Quality Assessment}
\begin{figure}[ht]
\centering
\includegraphics[width=0.6\linewidth]{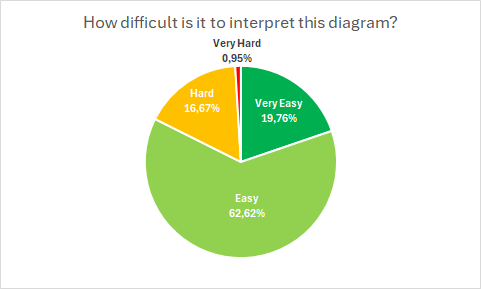}
\caption{}
\label{fig:diagram_quality_stats4}
\end{figure}

\begin{itemize}
\item \textbf{Is this a diagram?} Yes/No (Figure~\ref{fig:diagram_quality_stats1}). Annotators agreement (Gwet AC1): 0.6769. 

\begin{figure}[ht]
\centering
\includegraphics[width=0.6\linewidth]{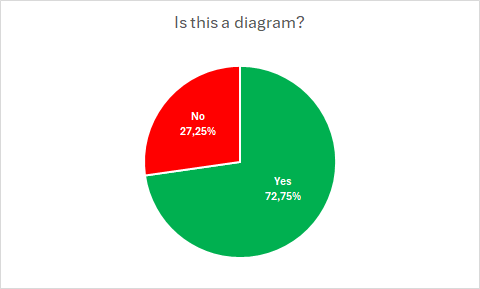}
\caption{}
\label{fig:diagram_quality_stats1}
\end{figure}

\item \textbf{Is this diagram suitable for a middle-school science textbook?} Yes/No (Figure~\ref{fig:diagram_quality_stats0}). Annotators agreement (Gwet AC1): 0.5194.

\begin{figure}[ht]
\centering
\includegraphics[width=0.6\linewidth]{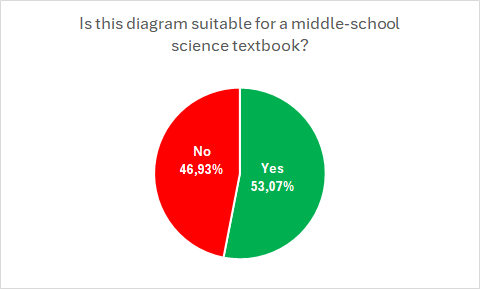}
\caption{}
\label{fig:diagram_quality_stats0}
\end{figure}

\item \textbf{How difficult is it to interpret this diagram?}
 Very easy/easy/hard/very hard (Figure~\ref{fig:diagram_quality_stats3}). Annotators agreement (Gwet AC1): 0,3714.

\begin{figure}[ht]
\centering
\includegraphics[width=0.6\linewidth]{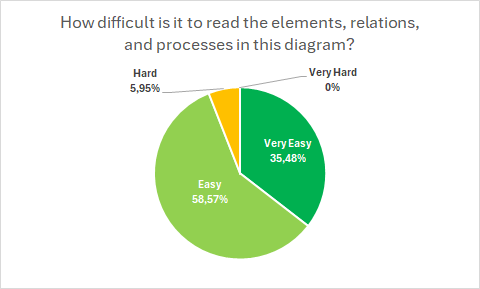}
\caption{}
\label{fig:diagram_quality_stats3}
\end{figure}

\item \textbf{How difficult is it to read the elements, relations, and processes in this diagram?} Very easy/easy/hard/very hard (Figure~\ref{fig:diagram_quality_stats4}). Annotators agreement (Gwet AC1): 0,3211.
\end{itemize}

\subsection{Caption Quality Assessment}
\begin{figure}[ht]
\centering
\includegraphics[width=0.6\linewidth]{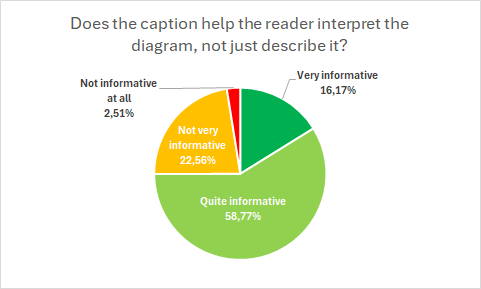}
\caption{}
\label{fig:caption_quality_stats5}
\end{figure}

\begin{itemize}
\item \textbf{Does the caption provide a coherent description of the diagram?} Very coherent/quite coherent/not too coherent/incoherent (Figure~\ref{fig:caption_quality_stats1}). Annotators agreement (Gwet AC1): 0,3535.

\begin{figure}[ht]
\centering
\includegraphics[width=0.6\linewidth]{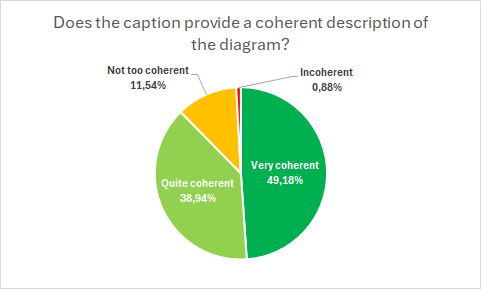}
\caption{}
\label{fig:caption_quality_stats1}
\end{figure}

\item \textbf{Does the caption cover all the elements, relations and processes involved in the diagram?} Yes, all of them/almost all of them; uncovered are not relevant/some of them are uncovered/most of them are uncovered (Figure~\ref{fig:caption_quality_stats2}). Annotators agreement (Gwet AC1): 0,8553.

\begin{figure}[ht]
\centering
\includegraphics[width=0.6\linewidth]{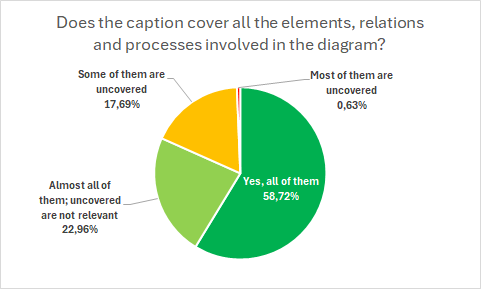}
\caption{}
\label{fig:caption_quality_stats2}
\end{figure}

\item \textbf{Are all the elements, relations and processes described in the caption present in the diagram?}
 Yes, all of them/some of them are not present/most of them are not present (Figure~\ref{fig:caption_quality_stats3}). Annotators agreement (Gwet AC1): 0,4336.
 
\begin{figure}[ht]
\centering
\includegraphics[width=0.6\linewidth]{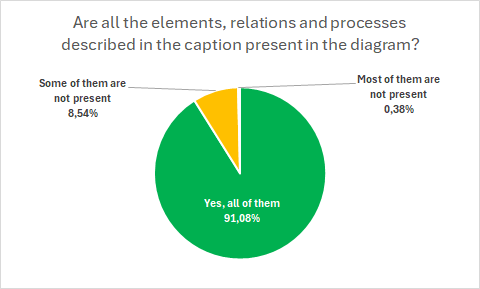}
\caption{}
\label{fig:caption_quality_stats3}
\end{figure}

\item \textbf{Does the caption use a clear, concise language at middle-school science level?} Very clear and adapted to the domain/quite clear and adapted to the domain/not very good adapted/not adapted at all (Figure~\ref{fig:caption_quality_stats4}). Annotators agreement (Gwet AC1): 0,2227.

\begin{figure}[ht]
\centering
\includegraphics[width=0.6\linewidth]{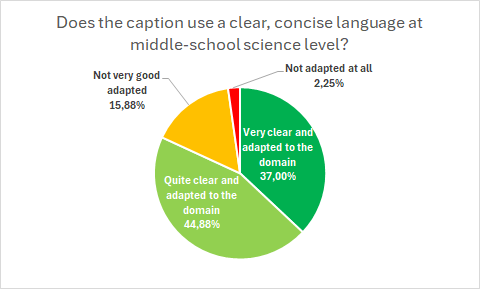}
\caption{}
\label{fig:caption_quality_stats4}
\end{figure}

\item \textbf{Does the caption help the reader interpret the diagram, not just describe it?} Very informative/quite informative/not very informative/not informative at all (Figure~\ref{fig:caption_quality_stats5}). Annotators agreement (Gwet AC1): 0,4004.
\end{itemize}

\subsection{MCQA Quality Assessment}
\begin{itemize}
\item \textbf{Is the question grounded in the diagram?} Yes/No (Figure~\ref{fig:mcqa_quality_stats1}). Annotators agreement (Gwet AC1): 0.8485.
\item \textbf{Can the question be answered without the diagram, based on commonsense or through a related text passage?} Yes/No (Figure~\ref{fig:mcqa_quality_stats1}). Annotators agreement (Gwet AC1): 0,5475.
\item \textbf{Is the question at the level of middle-school?} Yes/No (Figure~\ref{fig:mcqa_quality_stats1}). Annotators agreement (Gwet AC1): 0,6466.
\item \textbf{Is the question wording precise and free from ambiguity?} Yes/No (Figure~\ref{fig:mcqa_quality_stats1}). Annotators agreement (Gwet AC1): 0.8658.
\item \textbf{Do distractors reflect common misconceptions or errors, without being misleading?} Yes/No (Figure~\ref{fig:mcqa_quality_stats1}). Annotators agreement (Gwet AC1): 0,784.
\item \textbf{Are the answer choices clearly distinct, without overlap?} Yes/No (Figure~\ref{fig:mcqa_quality_stats1}). Annotators agreement (Gwet AC1): 0,8568.
\item \textbf{Is the correct answer actually correct?} Yes/No (Figure~\ref{fig:mcqa_quality_stats1}). Annotators agreement (Gwet AC1): 0,787.

\begin{figure*}[ht]
\centering
\includegraphics[width=\linewidth]{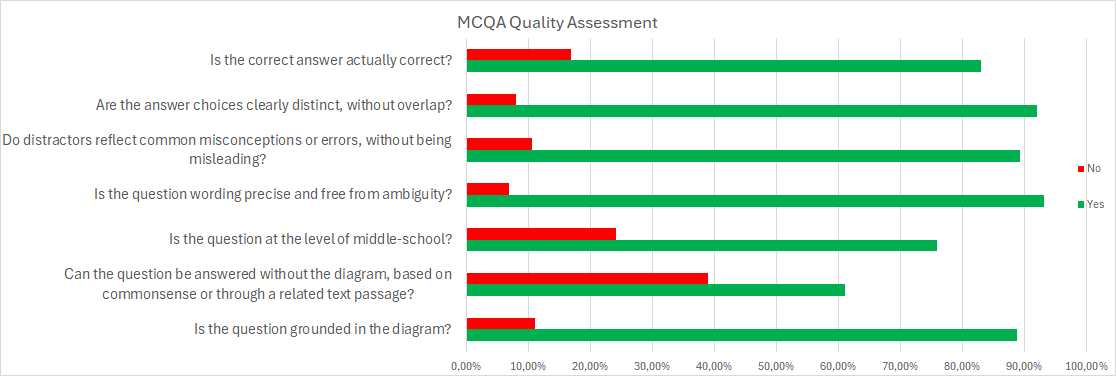}
\caption{}
\label{fig:mcqa_quality_stats1}
\end{figure*}
\end{itemize}

\section{Training Hyperparameters}
\label{sec:hyperparameters}
Table~\ref{table:hyperparameters} shows the hyperparameters used to train LLaVA-SciGram OV in each step of the pipeline.
\begin{table*}[ht]
\small
\centering
\resizebox{\textwidth}{!}{ 
\begin{tabular}{lp{5cm}p{6cm}p{6cm}}
\toprule
Hyperparameter & Alignment (SciGram-Align) & Visual Instruction Tuning (SciGram-VIT) & Further finetuning (SciGram-M\textsuperscript{3})\\ \midrule
Epochs & 1 & 1 & 3 \\ 
Lora R & - & 128 & 128 \\ 
Lora Alpha & - & 256 & 256 \\ 
Vision Tower & google/siglip-so400m-patch14-384 & google/siglip-so400m-patch14-384 & google/siglip-so400m-patch14-384 \\
mm tunable parts & mm mlp adapter & - & - \\
mm projector type & mlp2x gelu & - & - \\
mm vision select layer & -2 & -2 & -2 \\
mm use im start end & False & False & False \\
mm use im patch token & False & False & False \\
group by modality length & - & True & True \\
image aspect ratio & - & anyres & anyres \\
mm projector lr & - & 2e-5 & 2e-5 \\ 
Image Grid Pinpoints & - & [(384, 768), (768, 384), (768, 768), (1152, 384), (384, 1152)] & [(384, 768), (768, 384), (768, 768), (1152, 384), (384, 1152)]\\
Batch Size & 4 & 1 & 1\\ 
Gradient acc. Steps & 4 & 16 & 16 \\ 
Learning rate & 1e-3 & 1e-5 & 1e-5\\ 
Weight Decay & 0. & 0. & 0.\\ 
Warmup Ratio & 0.03 & 0.03 & 0.03 \\
Scheduler Rate Type & cosine & cosine & cosine \\
Model Max Length & 32768 & 32768 & 32768\\
Attn Implementation & sdpa & sdpa & sdpa\\
\bottomrule
\end{tabular}
}
\caption{Hyperparameters of SciGram stages.}
\label{table:hyperparameters}
\end{table*}

\section{Evaluation details}
\label{sec:evaluation_details}
During our evaluation on TQA, ScienceQA, and AI2D, we primarily used reported results from the literature for each model. However, some models did not have official results available, so we evaluated them ourselves using our custom prompts (Appendix~\ref{sec:qa_prompts}). The following list indicates which models were evaluated with our prompts and which relied on literature results:

\begin{itemize}
    \item \textbf{MemN+VQA} (TQA): literature results.
    \item \textbf{MemN+DPG} (TQA): literature results.
    \item \textbf{BiDAF+DPG} (TQA): literature results.
    \item \textbf{FCC+Vecsigrafo} (TQA): literature results.
    \item \textbf{IGMN} (TQA): literature results.
    \item \textbf{f-GCN1+SSOC} (TQA): literature results.
    \item \textbf{ISAAQ} (TQA): literature results.
    \item \textbf{Phi-3 Vision} (TQA, SQA, AI2D): our prompts for TQA, SQA, and AI2D Opaque; literature for AI2D Transparent.
    \item \textbf{MOLMo 7B-D} (TQA, SQA, AI2D): our prompts for TQA and SQA; literature for AI2D.
    \item \textbf{Pixtral 12B} (TQA, SQA, AI2D): our prompts for TQA, SQA, and AI2D Opaque; literature for AI2D Transparent.
    \item \textbf{Qwen2-VL 7B} (TQA, SQA, AI2D): our prompts for TQA, SQA, and AI2D Opaque; literature for AI2D Transparent.
    \item \textbf{Gemini 2.0 Flash} (TQA, SQA, AI2D): evaluated with our prompts.
    \item \textbf{GPT4o} (TQA, SQA, AI2D): our prompts for TQA, SQA, and AI2D Opaque; literature for AI2D Transparent.
    \item \textbf{LLaVa 1.5} (TQA): evaluated with our prompts.
    \item \textbf{LLaVA OV 7B} (TQA, SQA, AI2D): our prompts for TQA, SQA, and AI2D Opaque; literature for AI2D Transparent.
    \item \textbf{MCAN} (SQA): literature results.
    \item \textbf{Top-Down} (SQA): literature results.
    \item \textbf{BAN} (SQA): literature results.
    \item \textbf{DFAF} (SQA): literature results.
    \item \textbf{ViLT} (SQA): literature results.
    \item \textbf{Patch-TRM} (SQA): literature results.
    \item \textbf{VisualBERT} (SQA): literature results.
    \item \textbf{UnifiedQA Base} (SQA): literature results.
    \item \textbf{GPT-4 w/CoT} (SQA): literature results.
    \item \textbf{LLaMA-Adapter} (SQA): literature results.
    \item \textbf{Chameleon} (SQA): literature results.
    \item \textbf{LaVIN-13B} (SQA): literature results.
    \item \textbf{KAM-CoT} (SQA): literature results.
    \item \textbf{T-SciQ} (SQA): literature results.
\end{itemize}

\end{document}